\RequirePackage{fix-cm}
\documentclass[smallextended]{svjour3}       
\smartqed  
\usepackage[square,numbers,sort&compress]{natbib}

\usepackage{hyperref}

\usepackage{csquotes}
\usepackage{xcolor}
\usepackage[colorinlistoftodos,prependcaption]{todonotes}
\usepackage[inline]{enumitem}
\usepackage{xspace}
\usepackage{subcaption}
\usepackage{longtable}
\usepackage{afterpage}
\usepackage{amssymb}
\usepackage[nolist,nohyperlinks,printonlyused]{acronym}
\usepackage{amsmath}
\usepackage{pifont}
\newcommand{\cmark}{\ding{51}}%
\newcommand{\xmark}{\text{\ding{55}}}
\usepackage{makecell}

\usepackage{float}
 \usepackage[usenames,dvipsnames]{pstricks}
\usepackage{pstricks-add}

 \usepackage{pst-grad} 
 \usepackage{pst-plot} 
 \usepackage[space]{grffile} 
 \usepackage{etoolbox} 
 \makeatletter 
 \patchcmd\Gread@eps{\@inputcheck#1 }{\@inputcheck"#1"\relax}{}{}
 \makeatother
\usepackage{tikz}
\usetikzlibrary{mindmap,trees}
\usepackage{verbatim}
\usepackage{xspace}
\usepackage{acronym} 
\usepackage{multicol}
\usepackage{float}
\usepackage{array}
\usepackage{amssymb}
\usepackage{pifont}
\usepackage[ruled,lined,noresetcount ]{algorithm2e}
\usepackage{algpseudocode}
\usepackage{mathtools}

\usepackage{float}

\usepackage{geometry}
\begin{document}

\title{Federated and Transfer Learning for Cancer Detection Based on Image Analysis}


\titlerunning{Federated and Transfer Learning for Cancer Detection}        

 \author{Amine Bechar \and Youssef Elmir \and Yassine Himeur \and Rafik Medjoudj \and Abbes Amira}

 \authorrunning{Bechar et al.} 

 \institute{A. Bechar \at
               Laboratoire LITAN École supérieure en Sciences et Technologies de l'Informatique et du Numérique, RN 75, 06300, Amizour, Bejaia, Algérie
 \and              
 Y. Elmir\at
 Laboratoire LITAN École supérieure en Sciences et Technologies de l'Informatique et du Numérique, RN 75, 06300, Amizour, Bejaia, Algérie; SGRE-Lab, University Tahri Mohammed of Bechar, 08000, Bechar, Algeria
 \and
 Y. Himeur 
 College of Engineering and Information Technology, University of Dubai, Dubai, UAE
 \and
 R. Medjoudj\at
 Laboratoire LITAN École supérieure en Sciences et Technologies de l'Informatique et du Numérique, RN 75, 06300, Amizour, Bejaia, Algérie
 \and
 A. Amira\at
 Department of Computer Science, University of Sharjah, Sharjah, UAE
 ; Institute of Artificial Intelligence, De Montfort University, Leicester, United Kingdom\\
 \email{bechar@estin.dz; elmir@estin.dz; yhimeur@ud.ac.ae; medjoudj@estin.dz; aamira@sharjah.ac.ae;}           
              \emph{Present address:} of F. Author  
 }

 \date{Received: date / Accepted: date}

\maketitle

\begin{abstract}
This review article discusses the roles of federated learning (FL) and transfer learning (TL) in cancer detection based on image analysis. These two strategies powered by machine learning have drawn a lot of attention due to their potential to increase the precision and effectiveness of cancer diagnosis in light of the growing importance of machine learning techniques in cancer detection. FL enables the training of machine learning models on data distributed across multiple sites without the need for centralized data sharing, while TL allows for the transfer of knowledge from one task to another. A comprehensive assessment of the two methods, including their strengths, and weaknesses is presented. Moving on, their applications in cancer detection are discussed, including potential directions for the future. Finally, this article offers a thorough description of the functions of TL and FL in image-based cancer detection. The authors also make insightful suggestions for additional study in this rapidly developing area.

\end{abstract}


\section{Introduction}  \label{intro}

\subsection{Preliminary}
Cancer detection (CD) remains a pivotal topic in the contemporary health sector, largely due to cancer being the world's second most lethal disease following cardiac conditions \cite{anukriti2019investigation}. The predominant cause of this affliction is unrestrained cell growth, which can generate malicious tumors and adversely affect surrounding cells \cite{shrivastava2020bone}. Prompt CD not only significantly increases survival rates but also minimizes reliance on the visual examination of medical imagery by healthcare professionals, thereby reducing human error \cite{hamza2023hybrid}. A vital component of CD involved discerning between benign and malignant tumors, as they each require distinct treatment strategies \cite{tahmooresi2018early}. In light of the considerable strides made in digital image processing and artificial intelligence, numerous tools and frameworks have been developed to improve disease classification, detection, and even prediction, including cancer \cite{bechar2023harnessing}. Machine learning (ML), in particular, has emerged as a highly promising method for CD, owing to its ability to analyze complex patterns and make accurate predictions \cite{pradhan2020medical}.

Domains within ML, notably Deep learning (DL), have emerged as potent forces within the healthcare sector \cite{farrelly2023current}. The remarkable evolution of DL has facilitated the analysis of vast amounts of medical data, such as medical imaging, making the work of radiologists more efficient and effective \cite{wu2017small}. DL aids in extracting crucial insights, patterns, and anomalies, thereby enhancing the ability to detect and predict potential future diseases. However, ML and DL come with a set of limitations. The foremost among these is the challenge of determining which model best suits a specific use case within the healthcare system, particularly in terms of delivering optimal results. Additionally, the complexity of model training often necessitates considerable computational resources \cite{iman2023review}. Another significant constraint pertains to the dependency on a large set of labeled training data. Access to patient medical data, such as medical imaging, is often restricted due to privacy concerns, leading to issues related to the size and comprehensiveness of the dataset \cite{wittkopp2021decentralized}. This has implications for the accuracy and diversity of the models developed.

Addressing the inherent limitations of ML applications. Techniques such as Federated Learning (FL) and Transfer Learning (TL) have been introduced \cite{himeur2023video,kheddar2023deep}. FL is a decentralized ML approach that allows multiple parties to collaboratively train a model without the necessity of sharing sensitive medical data, ensuring data security and privacy \cite{bousbiat2023crossing}. On the other hand, TL, a subset of ML, facilitates knowledge transfer from multiple related datasets, thereby reducing an ML model's dependency on labeled training sets \cite{razavi2022introduction}. Nevertheless, training a model with limited datasets remains a challenging task. TL also aids in conserving time and resources during the model training phase. Therefore, a combination of FL and TL enables comprehensive learning from medical data while preserving user privacy. 

\subsection{Brief overview of CD using image analysis}
According to research by the World Health Organization (WHO) \cite{ferlay2021cancer}, cancer is a major cause of death worldwide, with predictions indicating that cancer rates in people are set to double in the near future. Early detection and treatment of cancer can significantly reduce the risk of mortality.
CD using image analysis involves the use of medical imaging techniques such as X-rays, computed tomography (CT), magnetic resonance imaging (MRI), and ultrasound to generate images of internal body structures. These images are then analyzed using various image analysis techniques to detect cancerous cells or tumors. Image analysis algorithms can be used to identify changes in the shape, size, and texture of cells and tissues that may be indicative of cancer. For example, computer aided detection (CAD) software can be used to analyze mammograms and detect early signs of breast cancer. Other techniques such as ML and DL can also be used to analyze medical images and identify patterns or anomalies that are indicative of cancer \cite{jeyaraj2019computer}. These algorithms can be trained on large datasets of medical images to improve their accuracy and sensitivity in detecting cancer \cite{jiang2023deep}.
\subsection{Importance of ML techniques in CD}

ML is the process of training a machine using a large amount of data in order to extract patterns and insights to make decisions \cite{ebert2016machine}. In recent years, ML has made significant improvements in the development of healthcare systems, especially in CD.

Cancer diagnosis and the healing process are assisted using ML, with supervised, unsupervised, and DL techniques (Neural Networks) \cite{saba2020recent}. Much research has been done using ML for various types of CD and diagnosis. For example, in \cite{xie2021early} the authors combined metabolomics and ML methods for early detection of lung cancer. In another study \cite{vaka2020breast} the authors used a deep neural network to improve the performance and the quality of breast cancer images.

In general, the methodology and steps of using ML algorithms in healthcare systems are the preprocessing, segmentation, feature extraction, training and classification, and performance evaluation \cite{mohammed2020analysis}. Recently, with the development of ML, many techniques and concepts have emerged to improve the performance of AI systems, such as FL and TL. These concepts address challenges in ML by improving accuracy and data privacy.
\subsection{Roles of FL and TL}

With the recent and significant advancements of healthcare systems and medical Internet of Things (IoT) devices, medical data has become available and easier to get and extract, this medical data aids in the detection of patient anomalies, and health issues and helps to monitor a patient's state.

Unfortunately, the availability of this data in centralized ML systems causes processing delays, and more crucially, it raises privacy concerns, which is a significant issue that has to be rectified. Another issue is that even if the data was safeguarded and made unavailable, it still presents a challenge. The issue of dataset limitations is what leads to overfitting problems if there are very limited labeled training datasets (difficult task to annotate medical data \cite{mehmood2021transfer}). As a result, the model's performance and quality of anomaly detection caused it to degrade and become unreliable.

To address these challenges, a multitude of concepts and methodologies have emerged in recent times. This study investigates FL and TL. The importance of TL is its powerful benefit in producing a medical model and using it in another identical model, which allows the problem of limited labeled training datasets and overfitting to be resolved. TL allows also improving the quality, precision, and stability of an anomaly detection model, and it reduces the computing rate and improve considerably the time of training a model in comparison with a model built from the ground up \cite{zheng2022application}.

On the other hand, the role of FL is to allow the training of ML models without the need to share the dataset. This is done using a federated server and federated clients instead of one centralized model \cite{alam2022federated}. The model was trained by each client using its datasets and sending the parameters and metrics back to the server through secured connections, this approach has drastically improved dataset security and integrity. 

Although FL decreases the accuracy of the model because of the different distribution of data for every FL client, TL resolves this issue by improving the quality and precision of the model to be ready to use \cite{weiss2016survey}.

\subsection{Paper's Contributions}
This review article delves into the utilization of FL and TL in CD through image analysis, highlighting their roles in enhancing ML applications in this domain. It offers a detailed examination of FL and TL, covering their definitions, advantages such as improved privacy, scalability, efficiency, and reduced training times, as well as their challenges like hardware limitations, communication issues, data distribution, and domain shift problems. The paper sets itself apart from existing literature by providing a comparative analysis of FL and TL, discussing their applicability across various types of cancer, and addressing specific considerations in choosing between the two methods. It also explores future directions and potential applications of these technologies in CD, while acknowledging the challenges related to data privacy, computational demands, and the heterogeneity of cancer. The review concludes with final thoughts and recommendations, aiming to guide future research in this rapidly evolving field. The main contributions of this review are summarized as follows:  
\begin{itemize}
    \item Providing the first comprehensive review of FL and TL for CD based on image analysis.
    \item Introducing a comprehensive taxonomy for FL and TL techniques for CD based on image analysis.
    \item Detailed comparison of FL and TL in terms of models, advantages, limitations, etc.
    \item Review of algorithms and models for applying FL and TL to CD.
    \item Identifying key challenges and future directions for these methods in CD.
\end{itemize}

\subsection{Comparison with other existing reviews}

This paper provides a comprehensive review of FL and TL techniques for CD based on medical image analysis. This paper is compared to existing survey articles along several key dimensions in Table \ref{tab:contr_comp}. The survey provides the most comprehensive coverage across FL, TL, model analysis, applications, and future outlook. Multiple cancer domains are spanned and the two techniques are examined. Other reviews usually only look at FL or TL separately. This review looks at how both methods work together to improve CD from medical images. A complete overview is provided of the roles and combined benefits of FL and TL. This makes this review unique and useful for researchers and healthcare professionals working in this field.

\subsection{Organization of the paper}
The remaining sections of the paper are structured as follows: Section 2 provides an overview of FL, including its definition, concept, advantages in CD, and challenges. Section 3 defines TL, and it discusses its advantages, limitations, and their applications in CD. Section 4 compares FL and TL based on image analysis, providing a taxonomy and models of each, along with a discussion of their advantages and disadvantages. Section 5 explores some applications and future directions of FL and TL for CD based on image analysis. Section 6 concludes with a summary of the key roles of FL and TL, thoughts on the future outlook for ML in CD, and final recommendations.

The paper aims to provide a comprehensive review of these two important ML techniques and their applications in improving CD based on medical image analysis. The organization begins by introducing each approach, then compares them, considers use cases, and looks toward the future.

\section{Federated Learning (FL)}

\subsection{Definition and concept of FL}
FL is a new approach of ML that allows a machine to train a model without sharing its sensitive data. The ultimate goal for each client is to share to a federated server weights, parameters, and even the metrics of the local model to make constant evaluation of the global model. The client-side established communications with the federated server and shared the weights, so it was aggregated to create a new global model \cite{konevcny2016federated}. This process happens and occurs in multiple rounds.

\color{black}
The fundamental objective in FL is to minimize a global objective function that is typically the weighted sum of local objective functions, formulated in \cite{yu2022survey}:
\begin{equation}
\min_{\theta} F(\theta) = \sum_{k=1}^{K} p_k F_k(\theta)
\end{equation}
where \(\theta\) represents the global model parameters to be learned, \(F_k(\theta)\) is the local objective function of the \(k\)-th participant (out of \(K\) total), and \(p_k\) is the weight of the \(k\)-th participant's local dataset, often chosen based on the dataset size or importance.

FL often uses gradient-based optimization methods, where the local updates can be computed using stochastic gradient descent (SGD) or its variants. Each participant computes the gradient of the local objective function with respect to the model parameters \cite{liu2020accelerating}:
\begin{equation}
\nabla F_k(\theta) = \frac{1}{N_k} \sum_{i=1}^{N_k} \nabla f(x_{k,i}, y_{k,i}; \theta)
\end{equation}
where \(N_k\) is the number of samples at the \(k\)-th participant, and \(x_{k,i}, y_{k,i}\) are the local data samples and their corresponding labels.

After computing local updates, these are sent to a central server (or aggregated in a decentralized manner) to update the global model. The simplest aggregation method is federated averaging (FedAvg), where the global model is updated as \cite{konevcny2016federated}:
\begin{equation}
\theta \leftarrow \theta - \eta \sum_{k=1}^{K} \frac{N_k}{N} \nabla F_k(\theta)
\end{equation}
with \(\eta\) being the learning rate and \(N\) the total number of samples across all participants.



FL can be applied to almost any edge device, this has revolutionized key fields such as medical healthcare applications \cite{li2023heterogeneity}. One of the key advantages of FL in medical systems is that it enables securing very sensitive and clinical medical data (CT, X-ray, MRI...) while maintaining the quality and performance of the model, as long as reducing execution time in comparison to centralized models \cite{liu2022decentralized}.

\begin{table}[t]
\centering

\caption{Contribution comparison of the proposed study against other FL and TL surveys. The tick mark (\cmark) indicates that the specific field has been addressed, whereas the cross mark (\xmark
) means addressing the specific fields has been missed }
\label{tab:contr_comp}

\begin{tabular}{llllllllll}
\hline
Review & Application & FL & TL  & Research & Public & Model  & Open &Future\\
& domain &   &  &  Questions & Dataset & Analysis &  Challenges &Direction\\  \hline
{\tiny \cite{meghana2023breast}} & Breast cancer & \xmark & \cmark &  \xmark & %
\xmark &\cmark & \xmark & \xmark \\ 
{\tiny\cite{rani2023application}} & Lung cancer & \xmark & \cmark & \xmark
&\xmark& \cmark &  \xmark & \xmark \\ 
{\tiny\cite{coelho2023survey}} & Healthcare & \cmark & \xmark &\cmark
&\cmark & \xmark &  \xmark & \xmark \\ 
{\tiny\cite{rahman2023federated}} & Healthcare & \cmark & \xmark & \xmark & %
\cmark &\xmark &  \cmark & \cmark \\ 
{\tiny\cite{chowdhury2021review}} & Healthcare & \cmark & \cmark &  %
\xmark & \cmark &\xmark &  \xmark & \cmark \\ 
{\tiny\cite{ayana2021transfer}} & Breast cancer & \xmark & \cmark &  \xmark & %
\cmark &\xmark &  \xmark & \cmark \\
{\tiny\cite{rauniyar2023federated}} & Healthcare & \cmark & \xmark &  \cmark &\xmark & %
\xmark & \cmark & \cmark \\ 
{\tiny\cite{hasan2022dermoexpert}} & Skin lesion & \cmark & \cmark &  \xmark &\cmark & \cmark &  \xmark & \cmark \\ 
{\tiny\cite{kumar2021federated}} & Healthcare & \cmark & \cmark &  %
\cmark &\cmark & \cmark &  \xmark & \cmark \\ 
{\tiny\cite{joshi2022federated}} & Healthcare & \cmark & \cmark & \xmark & \xmark &\cmark
&  \cmark & \cmark \\ 
Ours & Multidisciplinary & \cmark & \cmark & \cmark & \cmark & \cmark & %
 \cmark & \cmark \\ \hline
\end{tabular}

\end{table}

\subsection{FL Types}

\subsubsection{Horizontal FL (HFL)}
In HFL, the objective is to learn a global model \( \theta \) by minimizing the sum of local loss functions computed on local datasets that share the same feature space but differ in samples. Mathematically, the objective can be formulated as \cite{huang2022fairness}:
\begin{equation}
\min_{\theta} F(\theta) = \sum_{k=1}^{K} \frac{N_k}{N} F_k(\theta)
\end{equation}
where \( F_k(\theta) \) is the local loss function of the \( k \)-th participant, \( N_k \) is the number of samples of the \( k \)-th participant, \( N \) is the total number of samples across all participants, and \( K \) is the total number of participants.

\subsubsection{Vertical FL (VFL)}
In VFL, participants aim to learn a global model by leveraging datasets that have the same sample space but different features. The goal is to optimize a global objective function that may involve joining features from different participants to predict a common target. The mathematical formulation involves a coordinated optimization problem where each participant contributes a different part of the feature vector \cite{zhu2021pivodl}:
\begin{equation}
\min_{\theta} F(\theta) = \sum_{k=1}^{K} F_k(\theta_k)
\end{equation}
with \( \theta = (\theta_1, \theta_2, ..., \theta_K) \) representing the parts of the model parameters corresponding to the features held by each participant.

\subsubsection{Federated Transfer Learning (FTL)}
FTL seeks to transfer knowledge from one domain to another. Mathematically, this involves optimizing local models on their respective datasets and then transferring some aspects of these models (e.g., model parameters, representations) to improve learning in another domain with possibly different feature and sample spaces \cite{liu2020secure}:
\begin{equation}
\min_{\theta_s, \theta_t} F_s(\theta_s) + F_t(\theta_t, \theta_s)
\end{equation}
where \( F_s \) and \( F_t \) are the source and target domain objective functions, respectively, and \( \theta_s \), \( \theta_t \) are the model parameters for the source and target domains.

\subsubsection{Decentralized FL (DFL)}
DFL optimizes the global model without a central server, relying on local updates and peer-to-peer communication. Each participant updates its model based on local data and then aggregates updates from neighbors, which can be mathematically represented as \cite{beltran2023decentralized}:
\begin{equation}
\theta_i^{(t+1)} = \theta_i^{(t)} - \eta \left( \nabla F_i(\theta_i^{(t)}) + \sum_{j \in \mathcal{N}_i} w_{ij} (\theta_j^{(t)} - \theta_i^{(t)}) \right)
\end{equation}
where \( \theta_i^{(t)} \) is the model parameters of the \( i \)-th participant at iteration \( t \), \( \eta \) is the learning rate, \( \nabla F_i \) is the gradient of the local loss function, and \( w_{ij} \) are the weights denoting the influence between the \( i \)-th participant and its neighbors \( \mathcal{N}_i \).

The key difference among these types of FL lies in the formulation of the objective function, data partitioning, and model parameter sharing strategies. HFL and VFL differ primarily in the structure of the data they are designed to work with, while FTL focuses on transferring knowledge between domains. DFL distinguishes itself by the decentralization of the model update and aggregation process.

\color{black}

\subsection{Advantages of using FL for CD based on image analysis}

The great evolution of IoT devices (edge devices and wearable sensors), has helped in improving medical monitoring and anomaly detection. This is done by generating real-time medical data such as biometrics, blood tests, and medical images. These types of datasets, especially the medical images, have generated a huge overflow of data which has caused performance and security issues. So for that FL has been inaugurated to resolve these issues and make other benefits and advantages such as:
Improving privacy, especially with the availability of enormous and sensitive medical data, it has become more and more easily accessed and attacked. Security and privacy have emerged as a focal and very point to address and improve, FL has introduced a new approach that allows hospitals and medical patient not to send or share their sensitive data. On the other hand, the client only shares weights to the server so that it will be aggregated \cite{yang2022review}. So the data has not been stored in one place and was not intercepted in the communication process, which is a huge boost to the security and privacy of the datasets.

Furthermore, FL approach needs the collection of datasets from various hospitals from different locations, the diversity of data allows for building a more robust model which improves the quality of the detection of anomalies and potential symptoms that lead to cancer. The diversity of the datasets allow also the reduction of the bias, this improved the prediction and the performance of the model \cite{darzidehkalani2022federated}.


In addition, one of the advantages of FL is that it improves drastically time execution of the training model and reduces the necessity of a very large space in the server (to save all the hospitals' data) because the clients' machines send only weights and parameters, also the federated server does not require having computational power because the training was run on different clients. The scalability of the FL server has not been affected by adding new hospitals or patients. Because the training was done on the client machines and shared only the data, this decentralization improved the scalability of FL which is a very important advantage \cite{diaz2023study}.

\section{Transfer Learning}
\subsection{Definition and concept of transfer learning}
TL is a new technique that allows utilizing knowledge and insights that is acquired and learned from data and use this knowledge it in different but related domains, this allows ML algorithms sharing their experience to improve the performance and reduce the dependency on labeled datasets \cite{zheng2022application}. This approach is very useful if there is difficulty in collecting labeled medical data, which is a problematic in healthcare department.

\begin{figure*}[t]
        \begin{center}
        \includegraphics[width=0.7\textwidth]{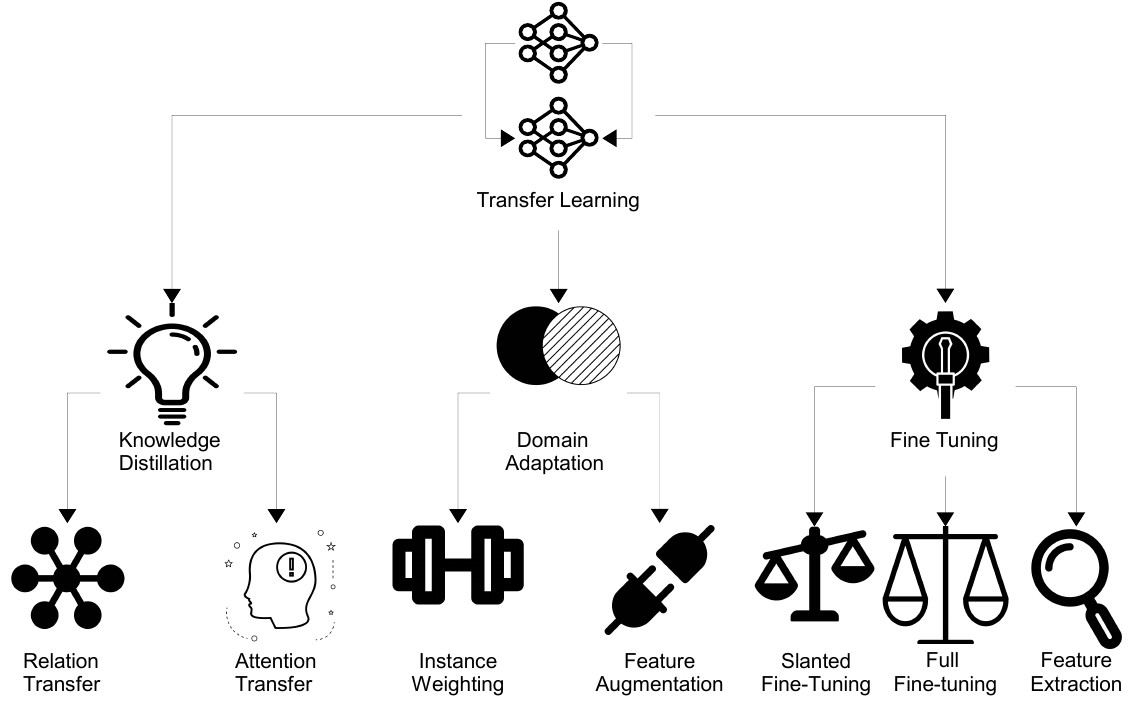}\\
        \end{center}
        \caption{A comprehensive examination of different types and subtypes of TL \cite{yang2019federated}}
        \label{fig:TTTL}
\end{figure*}

\color{black}
TL aims to improve learning in a new target task through the transfer of knowledge from a related source task. It involves two domains: the source domain $D_S$ with data distribution $P(X_S, Y_S)$ and the target domain $D_T$ with data distribution $P(X_T, Y_T)$. The goal is to use knowledge from $D_S$ to improve the learning of the target predictive function $f_T(\cdot)$ in $D_T$.
Typically, TL uses observations from multiple source domains and tasks to learn a decision function that can be applied to the target domain and task, with the aim of increasing diversity and improving the performance of the target class \cite{zhuang2020comprehensive}.
This is doable by minimizing the loss on the target domain, $L_T$, using knowledge from the source domain. Fig. \ref{fig:TTTL} provides a visual representation of the types and the subtypes of TL methods. 

\subsubsection{Fine-tuning}
Fine-tuning adjusts a pre-trained model to perform a new target task, involving minor modifications to the model architecture and re-training the model on the target task data.

\begin{equation}
    \theta^* = \arg\min_\theta L_T(\theta; D_T) + \lambda R(\theta)
\end{equation}
where $L_T$ is the target domain loss, $R(\theta)$ is a regularization term, and $\lambda$ is a regularization parameter.

\subsubsection{Domain Adaptation}
Domain adaptation is a technique in machine learning that aims to enable a model trained on a source domain to perform well on a different but related target domain, especially when there is abundant labeled data in the source domain and limited or no labeled data in the target domain \cite{wang2018deep}.
Domain adaptation seeks to transfer knowledge from a well-labeled source domain to a less-labeled or unlabeled target domain by minimizing both the domain discrepancy and the predictive loss on the target domain, using techniques such as discrepancy measures, adversarial training, and covariate shift adjustment.
Domain adaptation aims at minimizing the domain discrepancy between $D_S$ and $D_T$ and the predictive loss on the target domain $D_T$ \cite{you2019universal}.

\begin{itemize}
\item {Discrepancy Measures:} A common approach is to use discrepancy measures such as the maximum mean discrepancy (MMD) defined as \cite{lee2019sliced}:
\begin{equation}
\text{MMD}^2(F, X_S, X_T) = \sup_{f \in F} \left( E_{x_s \sim P(X_S)}[f(x_s)] - E_{x_t \sim P(X_T)}[f(x_t)] \right)^2
\end{equation}
where $F$ is a class of functions, and $E$ denotes the expectation.

\item{Adversarial Training:} Adversarial training can be formulated as a min-max problem \cite{su2020active}:
\begin{equation}
\min_{\theta_f} \max_{\theta_d} \left( E_{x_s \sim P(X_S)}[\log D(x_s)] + E_{x_t \sim P(X_T)}[\log(1 - D(x_t))] \right)
\end{equation}
where $D$ is the domain classifier, $\theta_f$ are the parameters of the feature extractor, and $\theta_d$ are the parameters of the domain classifier.

\item{Covariate Shift Adjustment:} Covariate shift adjustment involves re-weighting the loss function on the source domain samples as \cite{zhou2021bayesian}:
\begin{equation}
w(x) = \frac{P(X_T = x)}{P(X_S = x)}
\end{equation}
and modifying the source domain loss function accordingly. 
\end{itemize}
The optimization problem typically involves minimizing a combined loss function that includes the predictive loss on the source domain, the domain discrepancy measure, and the predictive loss on the target domain (if labeled data is available).

\subsubsection{Knowledge Distillation}
Knowledge distillation transfers knowledge from a complex model (teacher) to a simpler model (student), aiming to retain much of the teacher's performance.

\begin{equation}
    L = (1-\alpha)L_{\text{hard}} + \alpha T^2 L_{\text{soft}}
\end{equation}
where $L_{\text{hard}}$ is the traditional loss (e.g., cross-entropy against true labels), $L_{\text{soft}}$ is the distillation loss (e.g., KL divergence between the teacher's and student's predictions), $\alpha$ is a hyperparameter balancing the two losses, and $T$ is the temperature scaling the logits before applying softmax.

Overall, TL provides a broad framework for leveraging knowledge from related tasks while Fine-Tuning focuses on adjustments to a pre-trained model for a new task. On the other hand, domain adaptation addresses the discrepancy between source and target domain distributions. Moving on, knowledge distillation focuses on transferring knowledge from a complex model to a simpler one.

\color{black}
\subsection{Advantages of using transfer learning for CD based on image analysis}
In medical image analysis, TL has emerged as one of the most important techniques to improve the quality and the performances of DL models, and it comes with other advantages such as:
TL saves time by using a pre-trained system to learn new data. Instead of training a model from the ground up, which can be time-consuming and computationally expensive, as TL allows reusing pre-trained models that have already learned useful features. Reducing training time is very important, especially because many hospitals possess limited resources to train these models. TL emerges as an efficient way to resolve this major issue \cite{zhu2022super}.

Moving on, TL can improve medical image analysis by using pre-trained models on large datasets like  convolutional neural network (CNN) models such as VGG16 and ResNet-50 to extract features from medical images and improve classification accuracy. It can also reduce the need for large amounts of labeled data, which can be time-consuming and expensive to obtain because it is a difficult task to find hospitals that possess labeled and clean datasets that help improve the quality of the prediction and help the model to extract insights and anomalies from this data. 

General applicability and diversity are crucial in medical image analysis since it is frequently challenging to collect extensive and diverse datasets that accurately reflect the spectrum of variability in actual clinical situations. TL can improve the generalization performance of models on new and unknown data by utilizing pre-trained models that have already learned useful features from substantial and varied datasets. Additionally, TL can help overcome issues with overfitting, which is essential in improving the diversity of the model and the performance \cite{morid2021scoping}.

\subsection{Comparison of FL and TL in CD based on image analysis}

FL and TL have complementary strengths and limitations, FL improves the scalability and efficiency of ML systems, protects privacy, and is adapted to different situations. However, to create a more robust model, datasets from different hospitals in different locations are required and there are hardware limitations, communication, scheduling, and data distribution issues. TL improves the accuracy and sensitivity of ML models in detecting cancer, reduces the dependence of ML models on labeled training sets, and improves the quality and precision of anomaly detection models based on clinical data. However, the challenge is to select the most suitable model for a specific healthcare use case. Finally, it is noted that FL and TL can be used together, where FL has enabled distributed private training while TL has improved performance through the transfer of knowledge and learned functions. The choice depends on several factors: specific use case, data availability, privacy requirements, problem similarity, and computational limitations.  Table \ref{tab:comparaison} shows the key differences between FL and TL based on several factors.

\section{Federated Learning vs Transfer Learning}
\subsection{Overview}
This section provides a visual taxonomy of the key components of federated and TL techniques for CD based on medical image analysis. Fig. \ref{fig:Taxonomy} provides a visual representation of the taxonomy that categorizes these methods into different sections to compare their role in improving cancer diagnosis while maintaining privacy. Table \ref{tab:tabladeseables} summarizes research studies using FL and TL for CD through image analysis. The studies are compared based on parameters such as cancer type, DL model, learning method, aggregation strategy, dataset and best performance.
\subsubsection{Classification (C)}
This classification provides a general overview of the different categories and techniques for TL and FL applied in CD tasks:

\begin{enumerate}[itemindent=20pt,labelsep=4pt,labelwidth=1em,leftmargin=2pt,label= \textbf{C\arabic*.}]
     \item \textbf{Classification of TL: }
          This section provides an overview of the different classes and types of TL approaches and techniques developed to address the challenges of knowledge transfer from one domain to another:
                \begin{itemize}
                  \item \textbf{Data homogeneity: }is very important in generating high quality and performant models. The similarity of data distributions between the source domain and the target domain allows increasing the efficiency of TL, especially in healthcare where the data is complex. These are types of data homogeneity:

                  \begin{itemize}
                      \item \textbf{Homogenous Data: }Homogeneous datasets can be beneficial for TL as they contain similar data from the same domain, which improved the efficiency and quality of the model. Chui et al.  \cite{chui2023facilitating} points out that homogeneous medical datasets may not be available or provide enough diversity to build robust models, which is why TL with heterogeneous datasets has become an emerging research initiative.
                      \item \textbf{Heterogeneous data: }The presence of dissimilarity in data from different domains poses a challenge for TL when using heterogeneous datasets. This dissimilarity can lead to negative transfer, where the performance of the target model decreases after TL. Many researches have been conducted to solve this problem, such as Chui et al. \cite{chui2023facilitating} proposed a generic incremental TL approach to address the challenges of TL with heterogeneous datasets. This approach can be useful for small datasets and for applications that may not have similar datasets.
                   \end{itemize}
                   
                    \begin{figure*}[t]
                        \begin{center}
                        \includegraphics[width=1\textwidth]{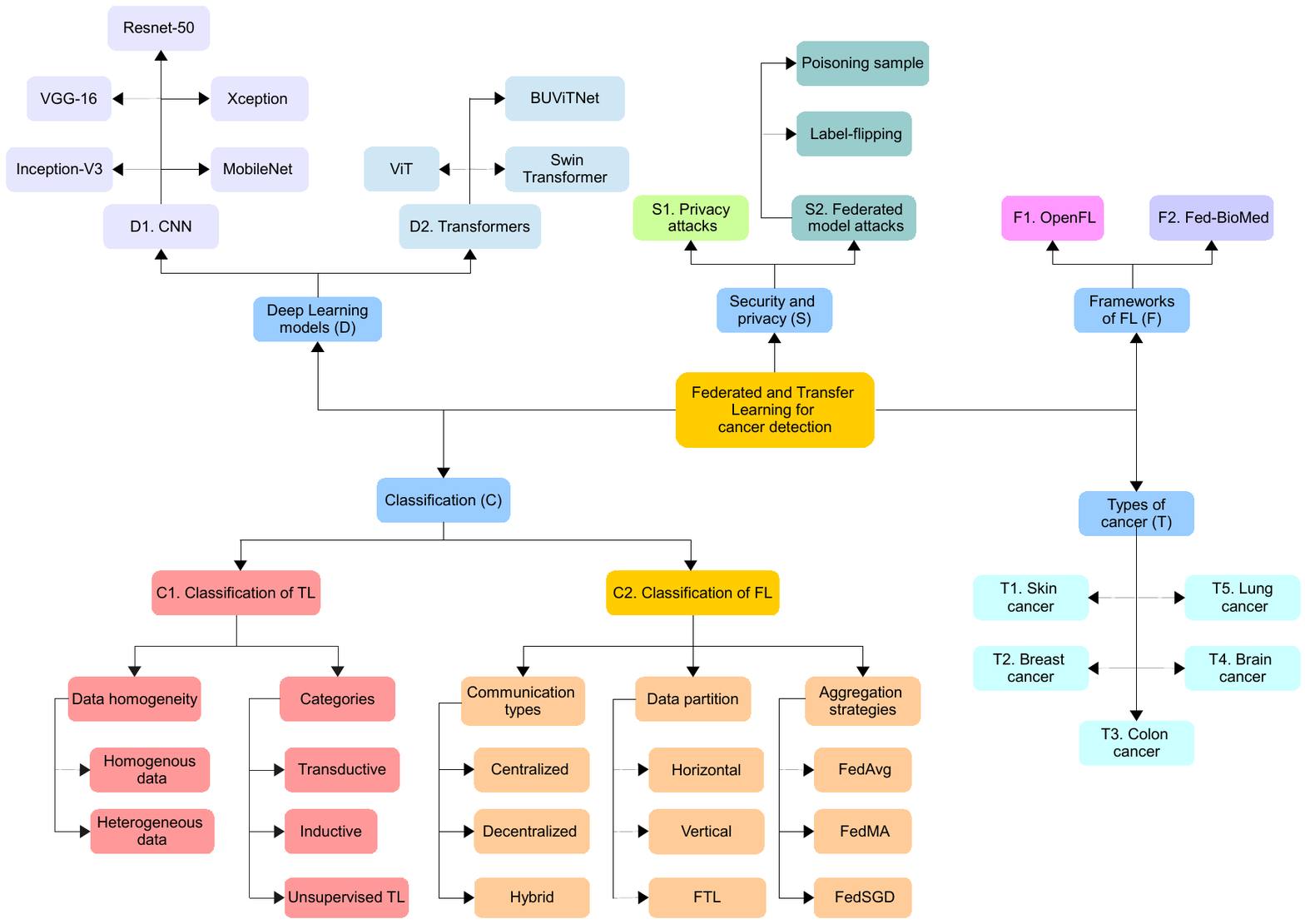}\\
                        \end{center}
                        \caption{Taxonomy of Federated Transfer Learning based on Image Analysis}
                        \label{fig:Taxonomy}
                    \end{figure*}

                    \begin{table*}
                        \centering
                        \caption{Key differences between FL and TL for CD}
                        \renewcommand{\arraystretch}{1.5}
                        \label{tab:comparaison}
                        \begin{tabular}{lp{6.3cm}p{6.3cm}}
                        \hline
                        \textbf{Key Differences} & \textbf{Federated Learning} & \textbf{Transfer Learning} \\ \hline
                        Technique & Decentralized ML approach that allows multiple parties to collaboratively train a model without sharing sensitive medical data, ensuring data security and privacy. &  Transfer knowledge from multiple related datasets, reducing an ML model's dependence on labeled training sets. \\ 
                        Data requirements & Requires datasets from various hospitals in different locations to build a more robust model. & Allows for the transfer of knowledge from one task to another. \\ 
                        Use cases & Used to improve the quality and performance of anomaly detection models while maintaining user privacy. & Used to improve the quality and precision of an anomaly detection model and reduce the computing resources required during the model training phase. \\ 
                        Performance & Improves the scalability and efficiency of ML systems. & Improves the accuracy and sensitivity of ML models in CD. \\ 
                        Privacy & Protect privacy by allowing hospitals and medical patients not to send or share their sensitive information. & Reduces the dependence of ML models on labeled training sets, thereby reducing the risk of privacy leaks.\\ 
                        Challenges & Hardware limitations, communication and scheduling problems, and data distribution. & Depends on a large set of labeled training data and determining which model is best suited for a specific healthcare use case. \\ \hline
                        \end{tabular}
                        \end{table*}
                    \begin{table*}
                        \centering
                        \caption{Summary of research studies conducted in FL and TL on CD based on image analysis}
                        \label{tab:tabladeseables}
                        \renewcommand{\arraystretch}{1.5}
                        \begin{tabular}{ccccccccc}   \hline
                        
                        \textbf{Ref} & \textbf{\makecell{Cancer \\ Type}} & \textbf{DL  Model} & \textbf{\makecell{Learning \\ Type}} & \textbf{\makecell{Method}} & \textbf{\makecell{Aggregation \\Strategy}} & \textbf{Dataset} & \textbf{\makecell{Best \\ Performance}}\\ \hline
                    
                        \cite{shamshiri2023compatible} & \makecell{Breast \\ } & \makecell{DenseNet-169} & TL & Inductive & N/A & BreakHis & \makecell{Accuracy=98.73\%} \\
                        
                        \makecell{\cite{rehman2020deep} } & \makecell{Brain \\ tumor} & \makecell{VGG16} & \makecell{TL} & \makecell{Inductive} & \makecell{N/A} & \makecell{Figshare brain \\ tumor} & \makecell{Accuracy=98.69\%} \\
                        
                        \makecell{\cite{talukder2023efficient}} & \makecell{Brain \\ tumor} & \makecell{ResNet50V2} & \makecell{TL} & \makecell{Inductive} & \makecell{N/A} & \makecell{Figshare brain \\ tumor} & \makecell{Accuracy=99.68\%} \\
                        
                        \makecell{\cite{lyu2022transformer}} & \makecell{Brain \\ metastases} & \makecell{Custom \\ transformer \\ model} & \makecell{TL} & \makecell{Inductive} & \makecell{N/A} & \makecell{In-house brain \\ MRI} & \makecell{AUC=0.878} \\

                        \cite{jimenez2023memory} & \makecell{Breast \\ } & CNN & FL & Vertical & FedAvg & 3 clinical datasets & \makecell{ROC= 5\%} \\

                        \cite{kumari2023magnification} & \makecell{Breast \\ } & \makecell{VGG-16, Xception,\\ Densenet-201} & TL & Inductive & N/A & \makecell{IDC, \\ BreaKHis} & \makecell{Accuracy=99.42\%} \\
                        
                        \cite{kumbhare2023federated} & \makecell{Breast \\ } & DenseNet, E-RNN & FL & Vertical & FedAvg & CBIS-DDSM & \makecell{Accuracy=95.73\%} \\
                        
                        \cite{talukder2022machine} & \makecell{Colon \\ } & YOLOv3 & TL & \makecell{Inductive} & \makecell{N/A} & \makecell{CVC, \\ colonDB} & Accuracy=96.04\% \\
                        
                        \cite{mehmood2022malignancy} & \makecell{Lung and \\ Colon} & \makecell{VGG16 and \\ VGG19} & \makecell{TL} & \makecell{Inductive} & N/A & LC25000 & \makecell{Accuracy=99.05\%} \\
                        
                        \cite{liu2023federated} & Lung  & ResNet18 & FL & Horizontal & FedAvg & Luna16 & Accuracy=83.4\% \\
                        
                        \cite{fang2018novel} & Lung  & GoogLeNet & TL & N/A & N/A & LIDC & Accuracy=81\%  \\
                        \cite{sajja2019lung} & Lung  & \makecell{Modified \\ GoogleNet} & TL & N/A & N/A & LIDC & Accuracy=99.03\%  \\
                        
                        \cite{heidari2023new} & Lung  & CapsNet & FL & Horizontal & \makecell{ Model \\ Aggregation} & \makecell{CIA, KDSB, \\ LUNA16, Local} & Accuracy=99.69\%  \\
                        
                        \cite{rajagopal2023federated} & Prostate & 3D UNet & FL  & Vertical & FedSGD & \makecell[l]{UCS, UCLA} & \makecell{IoU improved 100\%} \\

                        \cite{wang2020federated}& Various & \makecell[l]{LeNet, VGG} & FL & Vertical & FedMA & \makecell{MNIST, CIFAR-10} & N/A \\
                        
                        \cite{wang2022transunet} & \makecell{Brain \\ tumor} & TransUNet & TL & \makecell{Inductive} & N/A & BraTS & \makecell{Dice=0.864} \\
                        
                        \cite{masood2023multi} & Lung & \makecell{Swin \\ Transformer} & TL & \makecell{Inductive} & N/A & \makecell{LIDC-IDRI and ILD} & \makecell{Dice=0.9672} \\
                        
                        \cite{hosny2018skin} & \makecell{Skin}  & AlexNet & TL & Inductive & \makecell{N/A} & \makecell{Ph2} & Accuracy=98.61\%  \\
                        
                        \cite{cai2021many} & Skin  & CNN & \makecell{FTL} & Horizontal & \makecell{ Model \\ Averaging} & ISIC 2018 & Accuracy=91\%  \\
                        
                        \cite{tyagi2023amalgamation} & \makecell{Lung \\ tumor} & \makecell{ViT and \\ CNN} & TL & \makecell{Inductive} &  N/A & \makecell{NSCLC-Radiomics,\\ Local hospital} & \makecell{Dice=0.7468} \\
                        
                        \cite{li2023vision} & \makecell{Cell \\ tumor} & \makecell{ViT} & TL & \makecell{Inductive} & N/A & \makecell{BreakHis and \\ PatchCamelyon} & \makecell{AUC=0.96} \\
                        
                         \hline

                        \end{tabular}
                    \end{table*}    
                  \item \textbf{Categories: }Consists of three main types (inductive, transductive and unsupervised). The type depends on the availability and type of data. The types are: 
                      \begin{itemize}
                          \item \textbf{Transductive:} The transductive approach refers to a type of TL in which the source and target tasks are similar but their domains are different \cite{himeur2022next}. The goal of transductive TL is to improve the performance of a specific task in a target domain through the transfer of knowledge acquired from a source domain. This approach estimates the probability distribution of the target dataset given the source dataset and uses this data to adjust the source dataset to improve the performance of the child model \cite{slim2022improving}.
                          \item \textbf{Inductive:} According to \cite{paya2022automatic}, inductive transfer learning (ITL) is used to transfer knowledge from a source dataset to a smaller target dataset to improve the prediction efficiency of the given dataset. ITL is generally used in prior knowledge (neural networks and probabilistic generative processes) to detect unseen classes or clusters that are not seen or recognized by the original model.
                          
                          \item \textbf{Unsupervised Transfer Learning (UTL): }UTL is similar to inductive TL, with the main difference being the lack of labeled data in both the source and target domains \cite{agarwal2021transfer}, the method involved weighting the subspace-aligned features from the source users based on their agreement with the target user and subspace alignment to adapt the features from the source domain to the target domain \cite{chen2022multi}. There are many types of UTL such as unsupervised domain adaptation \cite{yang2023unsupervised}, Unsupervised domain generalization \cite{zhou2022asymmetrical} and Unsupervised Clustering \cite{wang2021survey}.
                      \end{itemize}
               \end{itemize}
                
               \begin{figure}[t]
                        \centering
                        
                        \begin{subfigure}{0.45\textwidth}
                          \centering
                          \includegraphics[width=0.85\textwidth]{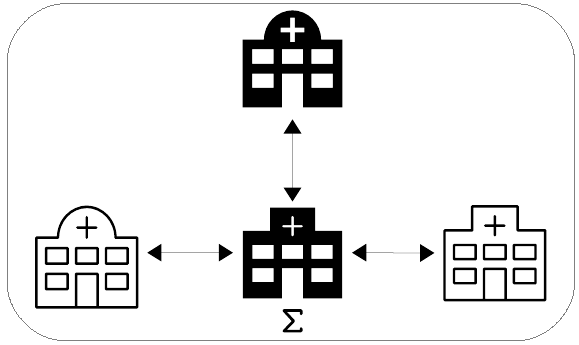}
                          \caption{Centralized FL with one aggregation server}
                        \end{subfigure}
                        \hspace{0.5cm} 
                        \begin{subfigure}{0.45\textwidth}
                          \centering
                          \includegraphics[width=0.85\textwidth]{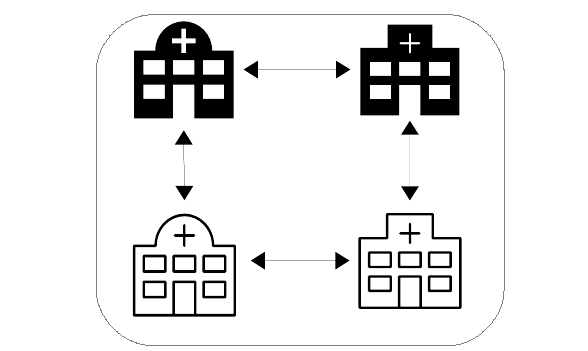}
                          \caption{Decentralized FL with one peer to peer communication}
                        \end{subfigure}
                        
                        \vspace{0.5cm}
                        
                        \begin{subfigure}{0.85\textwidth}
                          \centering
                          \includegraphics[width=\textwidth]{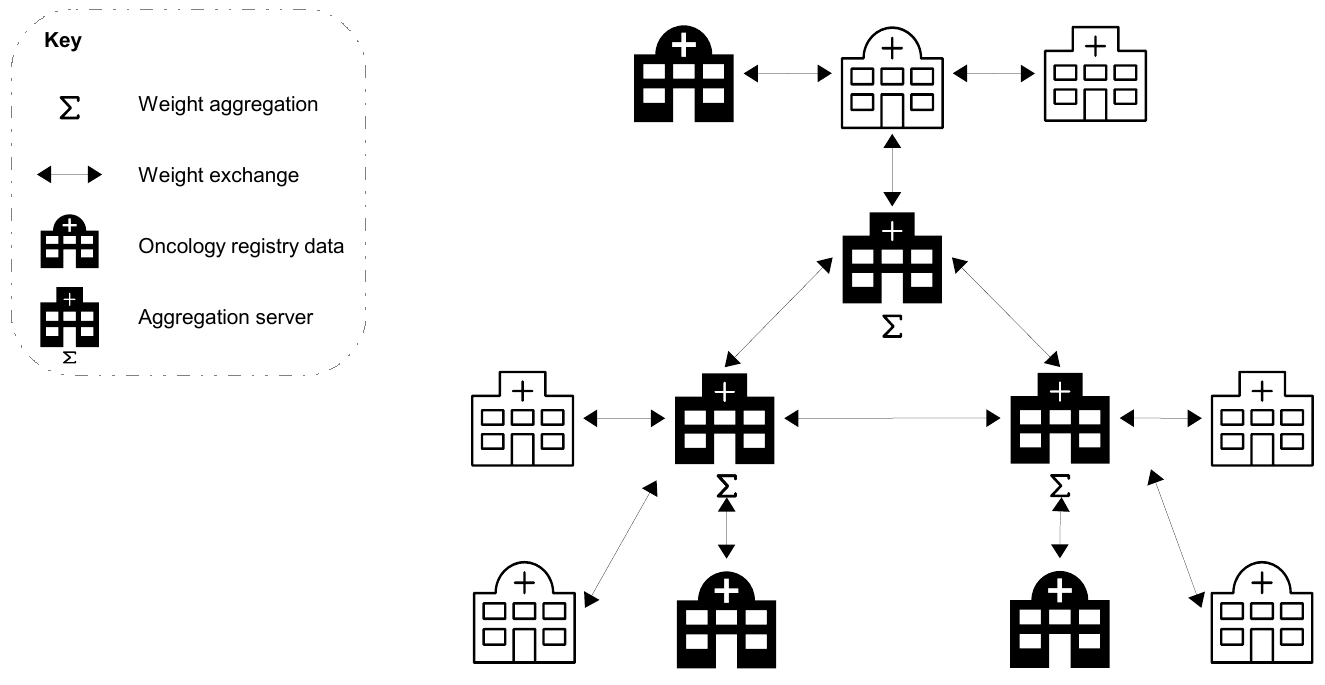}
                          \caption{Hybrid FL with three centralized FL connected with decentralized communication} 
                        \end{subfigure}
                    
                    \caption{Difference between centralized, decentralized and hybrid communication in FL environment for CD based on image analysis \cite{roy2019braintorrent,zhang2020hybrid}}
                        \label{fig:CDH}
                \end{figure}
          \item \textbf{Classification of FL: }This section focuses on the communication types, data partitioning and aggregation strategies used in FL. By providing an overview of the different techniques used in each of these categories, including their advantages and disadvantages:
                \begin{itemize}
                  \item \textbf{Communication Types: }Many types of ML models have been used in federated architecture to process and train medical data to extract insights, anomalies, and patterns.  Fig. \ref{fig:CDH} illustrates the three main communication styles, including centralized, decentralized and hybrid. These are the most commonly used ML types and models:
                    \begin{itemize}
                          \item \textbf{Centralized Communication: }This is the most common type of FL communication. It refers to a classic approach of using a central server to carry out operations and coordinate communications in a federated architecture, for example: planning the agenda of training rounds, aggregation strategy of weights, exchanging parameters with local customers and selecting customers, who took part in the training. The disadvantage of this communication is that the architecture is dependent on the server. If the server crashes for any reason, then the system collapses. Additionally, the number of maximum clients may be limited, leading to scalability issues. Therefore, decentralized communication was introduced to solve these problems \cite{abdulrahman2020survey}.
                          \item \textbf{Decentralized Communication:} refers to a type of communication in which two clients share model weight updates directly in a peer-to-peer communication protocol. The sharing of the weights covered all clients present in the federated network until the general model was formed, which allowed increasing the scalability of the federated model and ensuring that the federated model does not collapse \cite{beltran2023decentralized,roy2019braintorrent}. There are different types of federated decentralized communication, such as: gossip-based protocols, peer-to-peer protocols and blockchain-based protocols.

                          \item \textbf{Hybrid Communication:}  is a combination of centralized and peer-to-peer communication in a federated architecture. It is designed to explore the benefits of both centralized and decentralized communications, e.g. the control of the processes and strategies of the centralized model as well as their benefit from the scalability and security of the decentralized approach. Guo et al. \cite{guo2022hybrid} proposed a new algorithm called Hybrid Local SGD (HL-SGD) that uses both device-to-device (D2D) and device-to-server (D2S) communication. The authors concluded that this approach has accelerated the FL. However, the disadvantage of hybrid communication lies in the complex conception and implementation of these types of communication, as it combines many complex and different communication protocols.
                      \end{itemize}
                      
                  \item \textbf{Data Partition:} is a key aspect of FL, and it depends on the feature and sample space of the data parties involved. There are three categories: horizontal, vertical and federated transfer learning. Here is a brief explanation of each category:

                  \begin{itemize}
                          \item \textbf{Horizontal: }is an approach to training ML models on distributed datasets. The idea is to split the dataset by rows and store it in each participating client. For example, in this case, 03 hospitals were created. Each hospital's data has the same feature space but with different patients (even the same patient can have different IDs). Google Keyboard used this type of learning as the first use case of FL, where the participating mobile phones have the same functions with different training data \cite{mammen2021federated}. The main advantage of this method is that it can help address privacy concerns by storing the data at local hospitals. This can also reduce the amount of data to be sent, which is beneficial for hospitals with little computing power. 

                          \item \textbf{Vertical:} is a type of data partition that enables collaborative ML on vertically partitioned data while protecting privacy \cite{gu2021privacy}. The goal is to train a model on the features of multiple parties' data without explicitly sharing the raw data of parties \cite{das2021multi}. VFL divides a neural network between different parties and a server. Each party trains its local models based on their respective functions, and the server coordinates communication between the parties to generate a global model. The parties exchange encrypted model updates between each other and the server aggregates the updates to generate a global model \cite{yang2019parallel}. The biggest challenges of vertical FL include: Security and privacy; system heterogeneity and selection of important participants \cite{wei2022vertical}.

                          An example of the vertical architecture is shown in Fig. \ref{fig:VTL}, where the data of cancer patients is divided vertically as the features are divided based on the patient's personal data, details of the cancer disease (breast, melanoma, etc.), medical images (MRI, CT scans...) and finally medical analysis of the patients (Anatomical pathology, blood sugar...)
                          
                          \item \textbf{Federated Transfer Learning:} is an approach in FL that uses TL mechanism for learning aggregation sequentially \cite{ambesange2023simulating}. Fig. \ref{fig:FTTL} represent a visual representation of FTL. This involved transferring knowledge from a pre-trained model to a new model trained on a distributed dataset. The model of the target domain client is built using the diagnostic knowledge of the source domain clients and uploaded to the central server \cite{wang2023federated}. The steps to realize FTL includes : simulation of distributed data; simulation of FTL mechanism; selection and pre-training of the proposed model;  local data processing for model training and aggregating model learning at the master node \cite{ambesange2023simulating}. There are many advantages of FTL, some of them are: facilitating knowledge transfer with data protection, simultaneously addressing collaborative training, data protection and domain movement problems \cite{li2023federated}.
                   \end{itemize}
                        
                        \begin{figure*}[t]
                                \begin{center}
                                \includegraphics[width=0.85\textwidth]{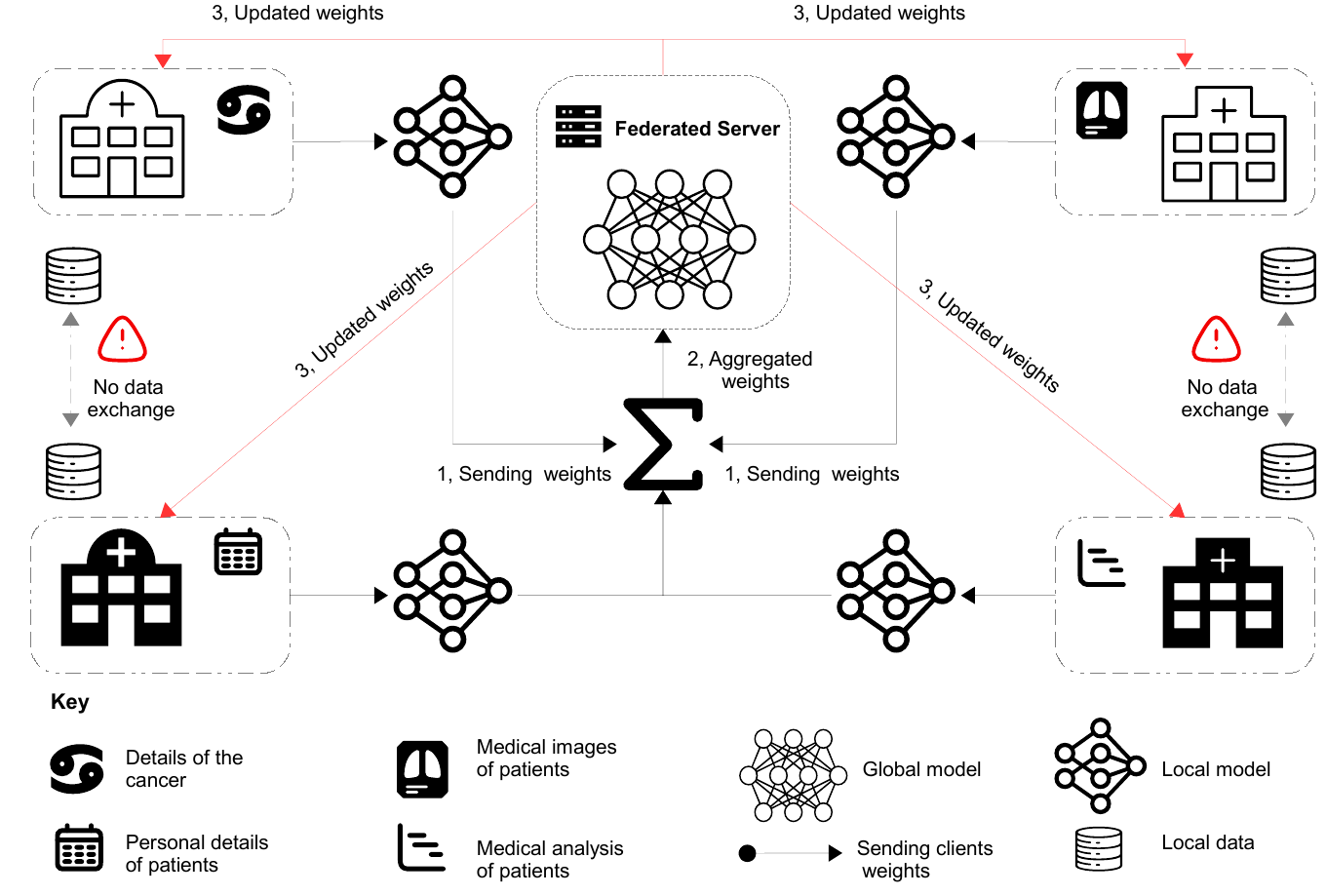}\\
                                \end{center}
                                \caption{The process of a round of FL based on a vertical partition architecture for cancer detection \cite{yang2019federated}}
                                \label{fig:VTL}
                        \end{figure*}
                        \begin{figure*}[t!]
                                \begin{center}
                                \includegraphics[width=\textwidth]{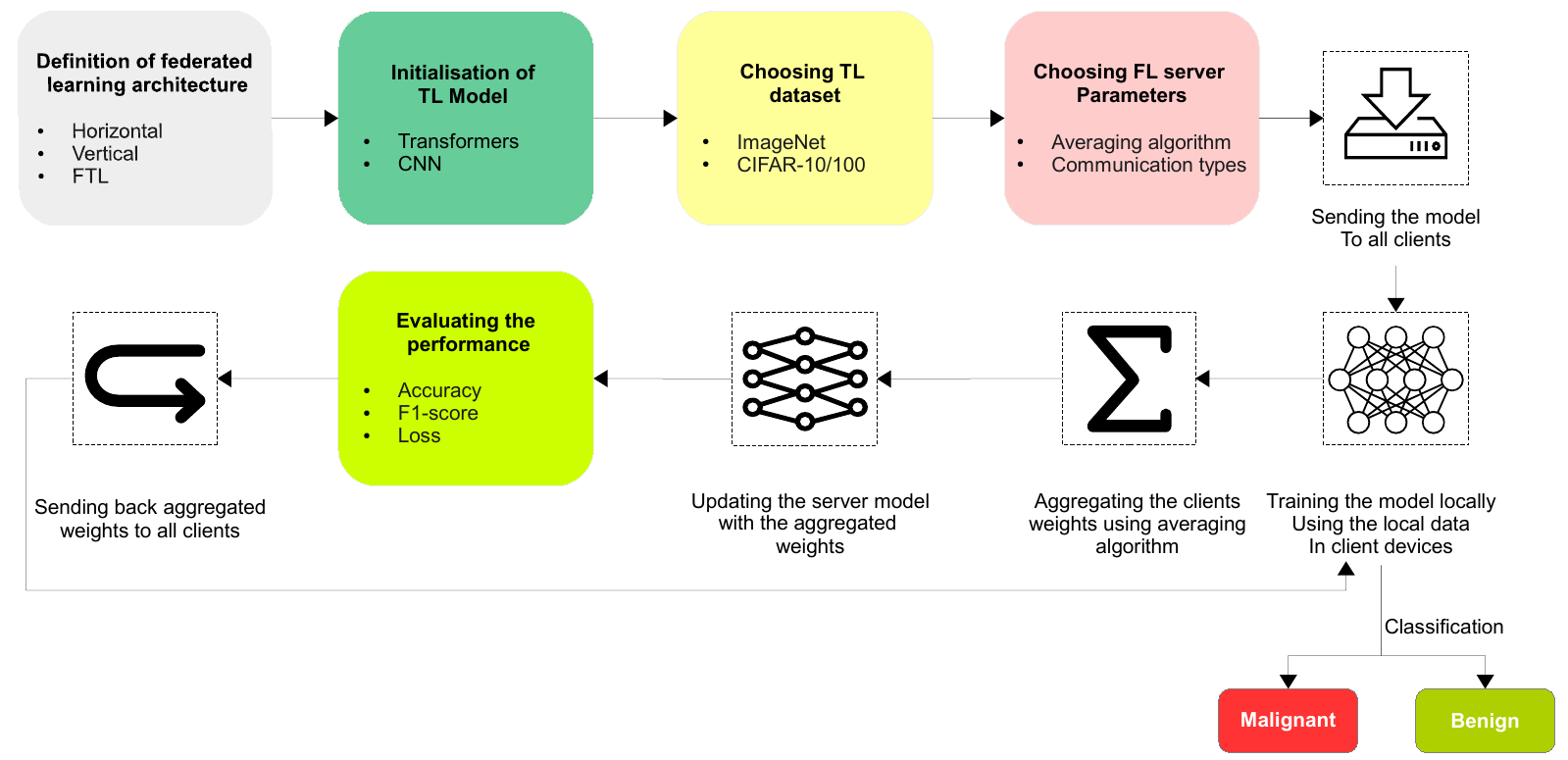}\\
                                \end{center}
                                \caption{The process of making a FTL for cancer detection}
                                \label{fig:FTTL}
                        \end{figure*}

               \item \textbf{Aggregation strategies:} are a crucial aspect of FL and involve combining different customers' local models to create a global model. There are different aggregation strategies in FL, some of them are:
                  \begin{itemize}
                    
                    \item \textbf{FedMA: }is a type of ML algorithm that helps build a common global model by matching and averaging hidden elements in each layer. This algorithm aims to address the heterogeneity of data and reduce communication congestion \cite{wang2020federated}. The FedMA approach aims to achieve device customization and high customer accuracy of one's own data while maintaining a high level of diversity \cite{ek2020evaluation}. The FedMA algorithm has several advantages over other FL approaches. Here are two key advantages: Efficient use of local models and addressing data bias \cite{wang2020federated}. One of the key limitations of FedMA is its reliance on well-trained local models. FedMA does not perform well in scenarios where customers have limited computing resources or insufficient training data. Another limitation is the limited applicability to certain neural network architectures (Long short term memory (LSTM) and CNN).   
                    
                    \item \textbf{FedSGD:} is a federated averaging algorithm that combines local Stochastic Gradient Descent (SGD) on each client with a server that performs model averaging \cite{mcmahan2017communication}. The algorithm selects clients and calculates the loss gradient over their local data using SGD. The gradients are then sent to the server where they are averaged to produce a global gradient update. This process is repeated over several rounds until convergence is achieved \cite{mcmahan2017communication}. Shin et al. \cite{shin2022fedvar} concluded that FedSGD works reasonably well with independent and identically d istributed (IID) or semi-IID data, but its performance degrades as data heterogeneity increases. Thonglek et al. \cite{thonglek2020federated} noted that FedSGD enables decentralized private training, but suffers from high communication costs and difficulties in dealing with heterogeneous devices/data. So, FedSGD is a federated averaging framework that aims to address the challenges of heterogeneity in federated networks.
                    \item \textbf{FedAvg: }is a synchronous-distributed optimization algorithm for FL to address the problem of communication efficiency in FL \cite{li2019convergence}.  FedAvg's process is to divide the training process into multiple rounds. In each round, the algorithm selects the clients who participated in the training. These clients then downloaded the model from the server and applied the training to their local data. Once training is finished, the weights and metrics are sent back to the server so that they are summarized in a global model. The same process is carried out in the new round of the algorithm until the maximum number of rounds is reached \cite{mcmahan2017communication}. This aggregation process averages the weights of the local models, giving each device or node an equal say in the final model. One of the main advantages of FedAvg is that it allows training ML models on distributed datasets without the need to transfer data to a central server. This helps address privacy concerns and can also reduce the amount of data that needs to be transferred. Fig. \ref{fig:fedavg} demonstrates the process of FL using FedAvg algorithm. Additionally, FedAvg addresses issues related to data heterogeneity \cite{darzidehkalani2022federated,casella2023benchmarking}. Fig. \ref{fig:Graph1} is the result of applying different aggregation strategies to REALWORLD datasets, where FedAvg performed the best \cite{ek2020evaluation}. However, one limitation of the FedAvg algorithm is that it can be affected by slow or unresponsive devices. Especially if there are many client devices with weak computing power \cite{gu2021fast}, this can affect the functionality of the algorithm. 
                \end{itemize}
               \end{itemize}
       \end{enumerate}
       
    \begin{figure*}[t!]
            \begin{center}
            \includegraphics[width=1\textwidth]{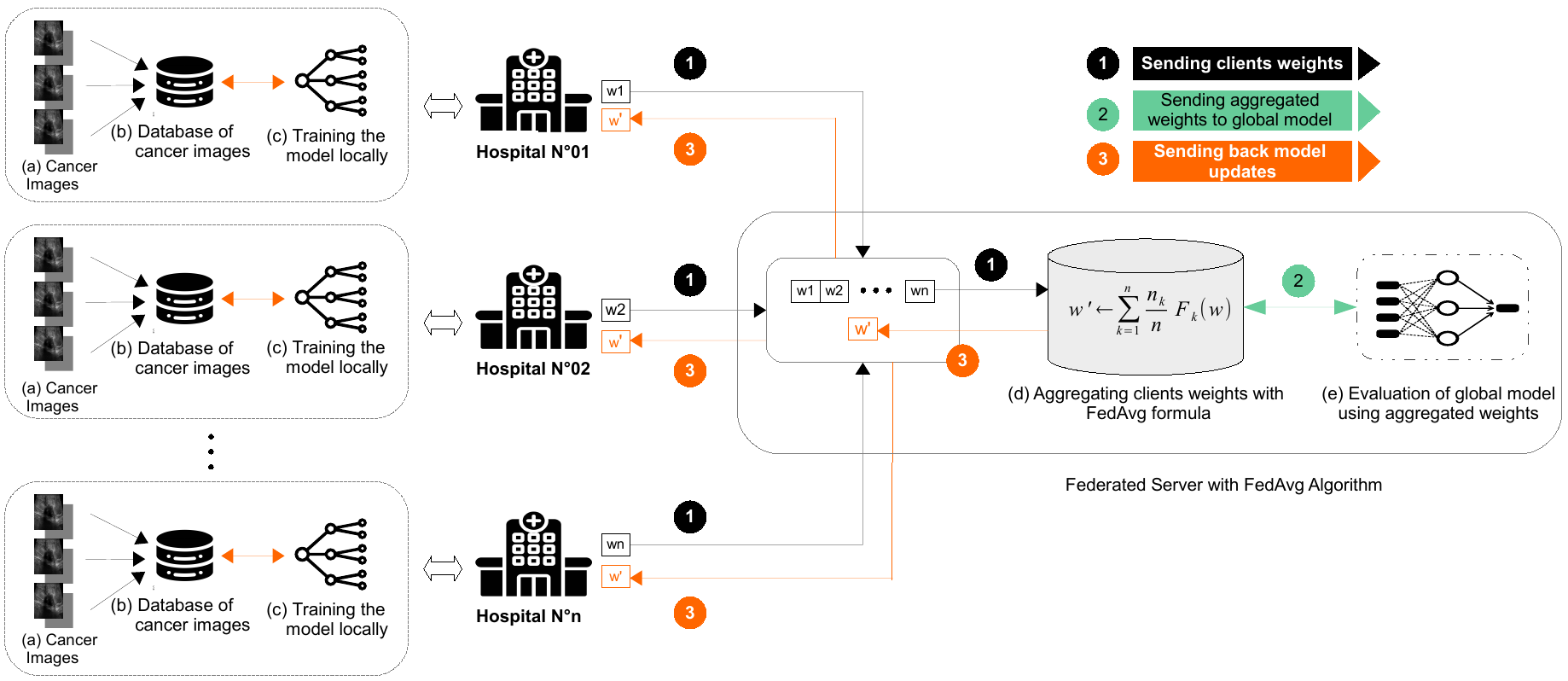}\\
            \end{center}
            \caption{The process of FL using the FedAvg averaging algorithm \cite{li2019convergence}}
            \label{fig:fedavg}
    \end{figure*}
    \begin{figure*}[t]
            \begin{center}
            \includegraphics[width=0.8\textwidth]{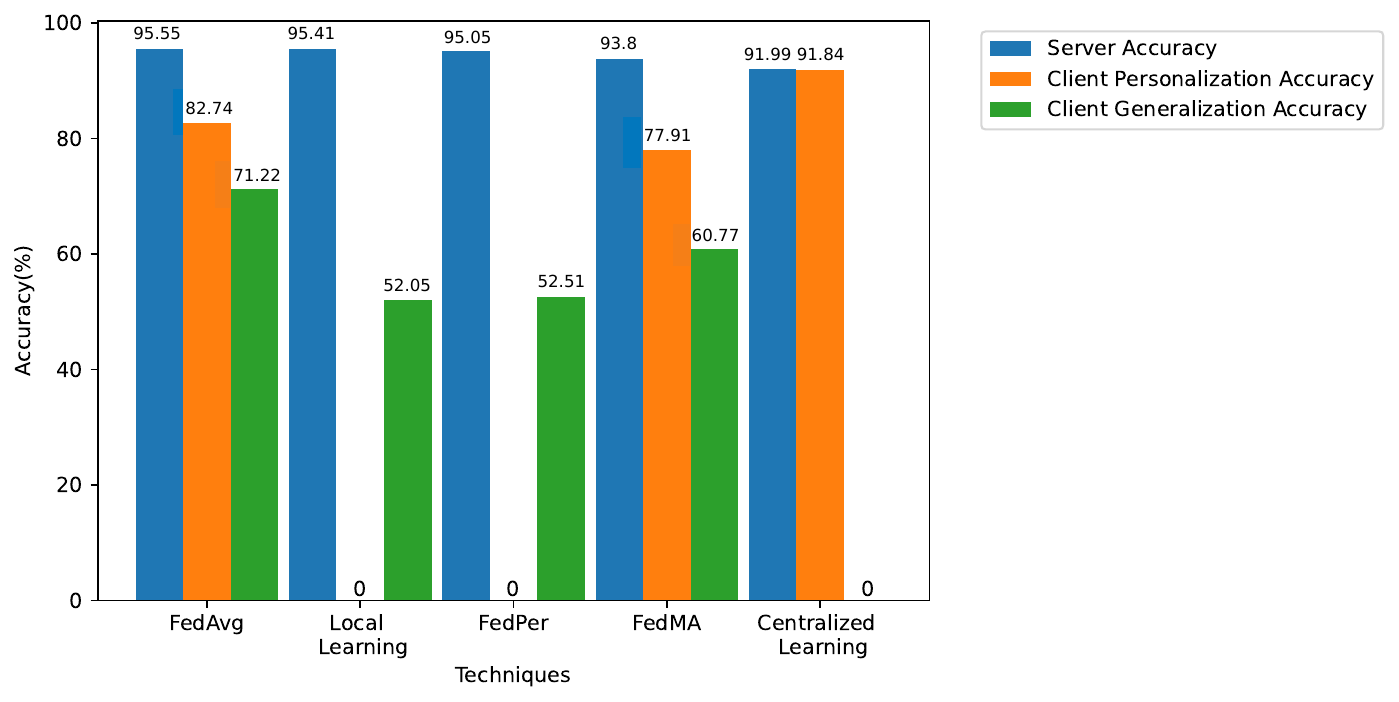}
            \end{center}
            \caption{Comparative study between centralized and FL accuracy results on the REALWORLD dataset \cite{ek2020evaluation}}
            \label{fig:Graph1}
        \end{figure*}

\subsubsection{Deep Learning models (D) } 
    
DL has become a game changer in computer vision, enabling computers to analyze and understand visual data with extreme accuracy. DL involves training neural networks on large datasets with a large number of features. There are two important categories of DL in computer vision:

\begin{enumerate}[itemindent=20pt,labelsep=4pt,labelwidth=1em,leftmargin=2pt,label= \textbf{D\arabic*.}]
    \item \textbf{Convolutional Neural Networks (CNN)}: CNNs are among the most commonly used neural network types in DL for images. These models are often used to teach computers to analyze images in a similar way to humans. This section examines the most widely used computer vision CNN models for medical imaging:
        \begin{itemize}
            \item \textbf{Inception-V3: }is a CNN model used for image classification tasks developed by Google researchers. Inception-V3 is based on the Inception architecture, which uses conventional layers having different filter sizes to extract image features (tumors for CD) \cite{ahmed2023inception}. Inception-V3 has been integrated in developing various DL models for CD. Li et al. \cite{liu2021deep} have built a model based on Inception-V3, R-CNN and S-Mask to improve the segmentation and classification of prostate ultrasound images, which helped in cancer diagnosis. Dong et al. \cite{dong2020inception} proposed a cell recognition algorithm that combined Inception-V3 and artificial features extraction for the classification of cervical cancer cell. This model provided effective methods for the diagnosis of the cervical cancer methods. Demir et al. \cite{demir2019early} used Inception-V3 and ResNet-101 for skin CD. The authors trained the models on 2437 training images, 600 test images and 200 validation images. The accuracy of the Inception-V3 model was 87\%. All this researches mentioned before showed that Inception model have given great results in CD based on image analysis, which ultimately improves the probability of successful therapy.
            
            \item \textbf{VGG-16: }is a pre-trained DL model used for medical image classification tasks. This architecture is composed of a total of sixteen layers, thirteen of which are convolutional layers and the remaining three are fully connected layers.  The function of the convolutional layers in a neural network is to extract characteristic from the input layers, while fully connected layers are used for classification \cite{sharma2023deep}. Various CNN models for detecting cancer have been developed using integration of VGG-16 models. Santos-Bustos et al. \cite{santos2022towards} presented an approach of uveal melanoma, a type of eye cancer. The two CNN architectures that were employed by the authors are VGG-16 and ResNet-18. The performance of the models have been assessed through various configurations and data augmentation techniques. The results indicated that the suggested method outperformed previous models obtaining 99\% accuracy and 98.4\% precision in identifying uveal melanoma. Although, the models were trained on a relatively small dataset (150 healthy and 33 unhealthy). Sharma et al. \cite{sharma2023deep} proposed a new method for pneumonia classification and prediction, the approach was to combine VGG-16 with Neural Networks. 
            VGG-16 was used in this article as a pre-trained feature extractor in the proposed method for pneumonia prediction in CXR images. The model can transfer the learned features from the large dataset of images so that the accuracy of pneumonia prediction can be improved. VGG-16 has also been used in detection of lung cancer, in \cite{pandian2022detection} the authors proposed an algorithm using pre-trained neural networks, specifically GoogleNet and VGG16 network for the purpose of classifying various types of lung cancer. These types include Adenocarcinoma, Large Cell Carcinoma, and Squamous Cell Carcinoma from normal lung images with an overall accuracy of 98\%. Fig. \ref{fig:VGG} shows the architecture and the role of each layers in VGG16. 
            
            \item \textbf{ResNet-50: }is a CNN architecture specifically designed for image recognition, particularly medical image analysis. ResNet-50 consists of five sets of 49 convolutional layers and a single fully connected layer. The convolution kernels used in this architecture have dimensions of 1x1, 3x3 and 5x5 and are used to extract image features at different resolutions \cite{wu2023improved}. The dimensionality of the output of the fully connected layer is directly related to the number of categories. Several CNN architectures have been developed to detect cancer by integrating ResNet-50 models. Sarwinda et al. \cite{sarwinda2021deep} proposed a method for detecting colorectal cancer using ResNet-18 and ResNet-50 models. The authors used a dataset of 165 images of the intestinal glands, consisting of 74 benign tumors and 91 malignant tumors. Therefore, based on the results presented in the article, the ResNet-50 model was the best model that gave the best results. Ikechukwu et al. \cite{ikechukwu2021resnet} conducted experiments using ResNet-50 as a pre-trained model to extract pneumonia images from normal chest X-ray images. The research showed that the validation accuracy of ResNet-50 was higher compared to other traditional techniques due to the effectiveness of using pre-trained models. B{\"u}t{\"u}n  et al. \cite{butun2023automatic} stated that ResNet-50 achieved 98.54\% classification accuracy in detecting cancer metastasis in lymph node images. However, the authors stated that the accuracy of ResNet architectures increases with increasing network depth, resulting in ResNet-101 providing the best results.
                
            \item \textbf{Xception: }is a deep CNN inspired by the Inception architecture. It represents a sequence of convolution layers characterized by depth separability and complemented by residual connections, which facilitates the definition and modification of the layers \cite{chollet2017xception}. Panthakkan et al. \cite{panthakkan2022concatenated} proposed X-R50, a hybrid model that combined the Xception and ResNet-50 networks. The role of the Xception architecture in this study was to capture fine-grained features that helped improve the results of accurate skin CD. Upasana et al. \cite{upasana2023attention} have developed a model for detecting pneumothorax in chest X-ray images by integrating the Xception neural network with an attention module. The proposed model was tested on 2,597 chest X-ray images and achieved a training accuracy of 99.18\%, a validation accuracy of 87.53\%, and an average AUC of 90.00\%. Sharma et al. \cite{sharma2022xception} proposed an approach using TL with the Xception model for feature extraction and implemented a support vector machine (SVM) with Radial Basis Function kernel to perform classification of histopathological images in the context of breast CD. The study showed that the proposed Xception+SVM R,5 model outperformed other competing DL models in terms of classification accuracy. Xception used depth-separable convolutions, which showed superior computational efficiency compared to traditional convolutions. Therefore, the algorithm should be made more efficient when handling and processing large-scale medical images.
                  
            \item \textbf{MobileNet: }is an architecture specifically designed for computing efficiency and optimized for development on mobile and embedded devices with limited computing resources. This was achieved by using depth-separable convolutions \cite{liu2021mmf}. Mothkur et al. \cite{mothkur2023classification} proposed a lung nodule classification approach based on deep hybrid learning to improve the accuracy of lung CD. The proposed approach leveraged MobileNet, which has proven that lung lesions with low memory requirements can be accurately classified as malignant or benign. Zhao et al. \cite{zhao2022machine} proposed a new method using the Mobilenet network structure for non-invasive classification of non-small cell lung cancer based on 18F-FDG PET/CT images. The MobileNet model showed good classification performance and can be used as a noninvasive technique to classify pathological subtypes of NSCLC. All of these studies have shown that MobileNet is efficient in classifying cancers without consuming important material resources.
        \end{itemize}
        \begin{figure*}[t]
                \begin{center}
                \includegraphics[width=\textwidth]{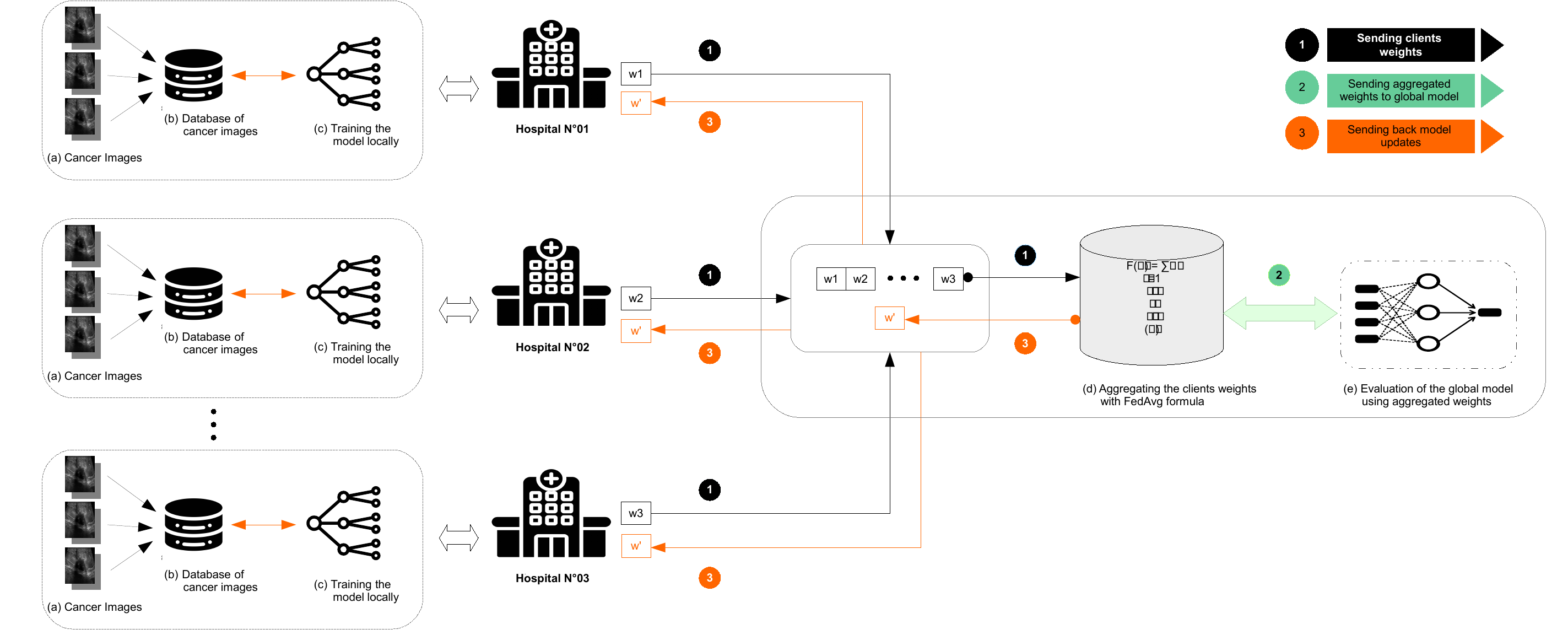}\\
                \end{center}
                \caption{VGG16 CNN architecture and layers for CD with two classes (malicious and benign) \cite{sharma2023deep,santos2022towards}}
                \label{fig:VGG}
             \end{figure*}  
        \item \textbf{Transformers: } are a type of neural network architecture based on self-attention mechanisms to model sequential data. They were first introduced for natural language processing tasks \cite{khan2022transformers}. There has been some recent work in the field of medical imaging using transformers to detect cancer in histopathological or radiological images \cite{he2023transformers}. This section lists the most commonly used transformer models for CD based on medical image analysis:
            \begin{itemize}
                \item {\textbf{Vision Transformers (ViTs):} }are transformative models that combine image analysis and self-attention-based architectures. They have revolutionized the field of computer vision and achieved excellent results in various tasks using the mechanism of self-attention. ViTs handle scalability well and are better than CNNs in terms of accuracy with more parameters, data and computing power. For example, ViT-G, a ViT model with 2B parameters, achieved an ImageNet accuracy of 90.45\%, outperforming CNNs \cite{zhai2022scaling}.

                \qquad In medical image analysis, an input image is divided into smaller patches, which are processed to obtain patch embeddings. Position encodings are added to extract spatial information. Patch embeddings are then fed to a multi-layer transformer encoder, each containing a multi-head self-attention (MSA) block and a multi-layer perceptron (MLP) block. The MSA block detects global dependencies between the image patches, while the MLP block enables the ViT model to learn patterns and improve the encoding of image patches to outputs. After several transformer-encoder layers, a global feature representation of the complete image is obtained and passed to the global encoder. The global encoder aggregates the encoded patch embeddings and models relationships between all patches. A sequence is output that represents the full image with global context. This sequence is then passed to the classifier head, which predicts a label for the medical image \cite{tyagi2023amalgamation,li2023vision,wang2023vision}. This entire process is explained in Fig. \ref{fig:VIT}.

                \qquad ViTs have been used in several CD studies, Li et al. \cite{li2023vision} proposed ViT-WSI, a ViT model for weakly supervised learning on whole slide images of brain tumor histopathology. For this study, the authors used multiple datasets such as internal brain tumor, TCGA and molecular biomarker datasets. ViT-WSI achieved a Macro AUC of 94.1\% on over 5,000 primary brain tumor slides, outperforming other methods such as CNN classifiers. Although the authors pointed out that the limitation of the approach is the high memory usage during attribution analysis. Andrade-Miranda et al. \cite{andrade2022pure} conducted a comparative study of various ViT models on BraTS2021 and found that the best performing model was the hybrid MCNN+ViTv-B/1 (combining CNN layers for feature extraction with Transformer-Layers for modeling long-range dependencies) with an average Dice of 91.1\% across 1251 3D MRI scans. The pure transformer models performed significantly worse than hybrid models with an average Dice of 87.3\%. Lu et al. \cite{xu2022brain} proposed a deep anchor attention learning (DAA) that uses a ViT and anchor-based attention to predict survival of brain tumor patients from MRI scans. DAA achieved C-index values of 0.69-0.70 when applied to 326 MRI scans from the BraTS 2020. Therefore, although the dataset size for DL methods is still relatively small, the C-index value is not high enough. In this study, ViTs performed better than using Resnet-18 for feature extraction.
                
                \qquad All of these studies conclude that ViTs are important for CD. The use of ViTs in CD has shown promising results and could be a valuable tool to improve cancer diagnosis.
                
                \item {\textbf{Swin Transformer:} }is a type of neural network architecture based on Transformers, suitable for processing images. It was created in 2021 by Microsoft Research Asia \cite{xie2021melanoma}. Swin Transformers leverage a window-based self-attention mechanism to reduce computational complexity and model cross-window relationships \cite{liu2021swin}. They have been successfully used in pre-training models for 3D medical image analysis and medical image segmentation tasks \cite{jiang2022swinbts,karthik2023dual}.  Swin Transformers bring the power of self-attention modeling to convolutional architectures like UNet to detect global context and long-range dependencies in images. To this end, many researchers have used Swin Transformers as feature extraction to improve medical image segmentation, such as Igbal et al. \cite{iqbal2023bts} proposed BTS-ST from breast tumor segmentation, Swin Transformer was used as an encoder to extract global context features from the input images. BTS-ST achieved an F1 score of 90.8\% in ultrasound and an F1 score of 85.6\% for classification in an ultrasound dataset from Shantou University Hospital. However, the dataset size was relatively small (42 patients). Zidan et al. \cite{zidan2023swincup} proposed SwinCup, a segmentation model that uses Swin Transformers as encoders in an encoder-decoder architecture for two colon histology datasets. SwinCup achieved an average F1 score of 90\% and an F1 score of 89\% on the CRAG dataset, although more experience with larger datasets is an advantage would have been. Masood et al. \cite{masood2023multi} proposed a modified Swin Transformer architecture ST-MSMLFFR for lung tumor segmentation in CT scans. The authors introduced a two-branch encoder with separate local and global feature extraction paths. ST-MSMLFFR achieved 96\% Dice on the LIDC-IDRI dataset, outperforming UNet (82\% Dice). Zou et al. \cite{zou2023swine} proposed a contribution to improve medical image segmentation by exploiting Swin Transformer's self-attention on the ISIC 2017 skin lesion dataset with 3029 CT scan slices from 108 cases. 88.47\% Dice and 92.29\% precision were achieved \cite{zou2023swine}. But the authors only trained their proposed model on two datasets, the performance of the model on other datasets was not discussed.
                \begin{figure*}[t]
                    \begin{center}
                    \includegraphics[width=\textwidth]{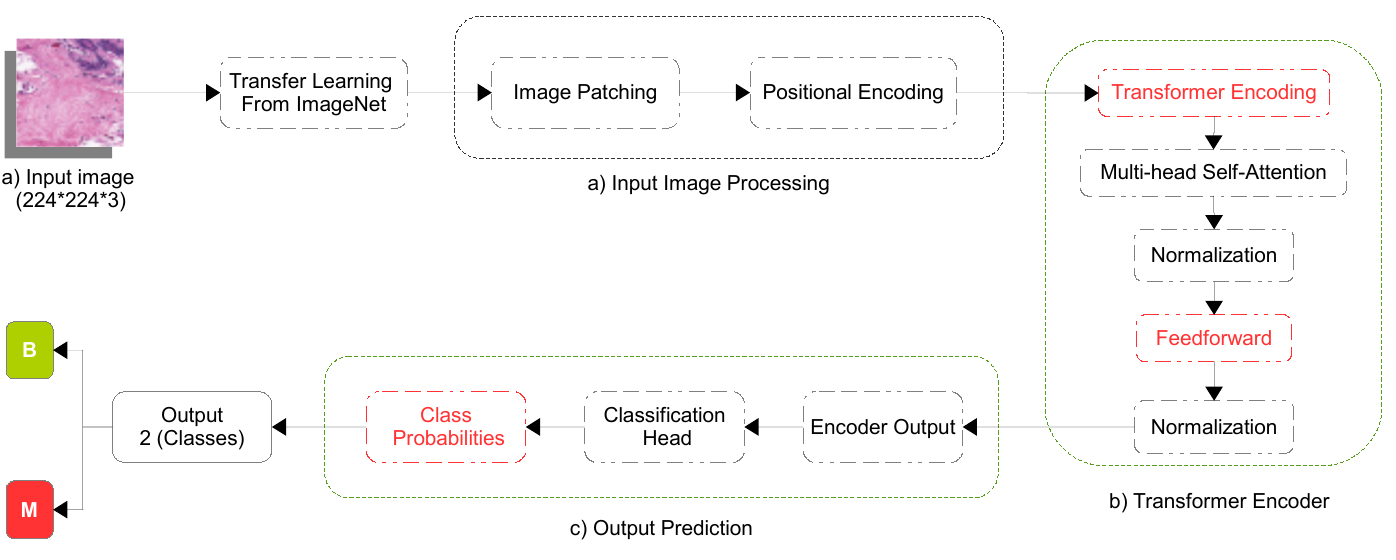}\\
                    \end{center}
                    \caption{ViT architecture and layers with TL for CD in two classes (malignant and benign)\cite{khan2022transformers,andrade2022pure}}
                    \label{fig:VIT}
                 \end{figure*}
                
                \item {\textbf{TransUNet:} } is a DL architecture for medical image segmentation that combines a transformer encoder with a U-Net decoder. It was developed by Chen et al. \cite{chen2021transunet} in 2021. The approach combined the ability of Transformers to globally extract the high-level relationships between image components with the precise localization ability of U-Nets. The encoder used a ViT to encode image patches and the decoder used U-Net to recover spatial details \cite{castro2022u}. Many researches have been conducted on CD. For example, Wang et al. \cite{wang2022accurate} used TransUNet to classify lung nodules on CT scans into benign and malignant categories. The method was evaluated on the LIDC-IDRI dataset of 8,474 CT images and achieved a specificity of 93.17\% and an AUC of 86.2\% on the test set, outperforming other methods such as stacked autoencoders and CNNs. However, the authors did not discuss limitations of the proposed TransUnit approach. Chen et al. \cite{chen2022brain} proposed an approach that combined TransUNet with a convolutional block attention module (CBAM). This enabled both local features and global contextual information to be captured for accurate segmentation. Experiments were conducted on the BraTS 2021 dataset, which contained multimodal MRI scans of brain tumor patients. The proposed TransUNet achieved Dice scores of 93.08\% for the whole tumor, 91.49\% for the tumor core, and 87.76\% for the tumor enhancement. This has outperformed other methods such as 3D UNet, Swin UNETR and VT-UNet. Wang et al. \cite{wang2022transunet} used TransUNet with a refined loss to achieve better segmentation performance of brain tumors on MRI images. In this approach, CBAM was used for the upsampling part of the U-Net architecture. This helped the model focus on the most important features in the images when upsampling. In experiments, TransUNet achieved a higher Dice similarity coefficient (86.4\%) compared to regular U-Net (81.1\%) and TransUNet without CBAM (65.3\%) in 3929 brain MRI cases. Kaggle datasets. Chen et al. \cite{chen2021mt} experimentally trained TransUNet on three public datasets - ISIC-2017, ISIC Additional and PH2 for skin lesion segmentation and classification, achieving a Dice score of 87\%. However, the authors stated that performance degrades when the images are obscured (e.g. by hair...).
                 
            \end{itemize}
        \end{enumerate}
\subsubsection{Frameworks of FL for ML Applications (F) }
    There are several open source frameworks available for implementing FL for ML applications that provide the necessary tools and infrastructure. Choosing the right framework depends on the specific use case and the requirements of the project. This comparison highlights the different characteristics and capabilities of each study's approach to FL and TL for CD, as well as the differences between current solutions in this area. Here are some popular frameworks for FL:
        \begin{enumerate}[itemindent=20pt,labelsep=4pt,labelwidth=1em,leftmargin=2pt,label= \textbf{F\arabic*.}]
              \item \textbf{OpenFL: }is an open source software library designed for FL applications. The library allows developers to build an ML model and train it on remote data owners or collaborating site nodes. The development of OpenFL is the result of a collaboration between Intel Labs and the prestigious University of Pennsylvania \cite{foley2022openfl}. By using OpenFL, companies can jointly train AI models while ensuring the protection of sensitive information and simplifying data sharing challenges. Reina et al. \cite{reina2021openfl} used OpenFL to train a consensus ML model used to identify and evaluate the boundaries of brain tumors. Foley et al. \cite{foley2022openfl} used OpenFL to ensure the privacy and security of a DL model to predict the likelihood of acute respiratory distress syndrome and death in COVID-19. All of these researchers concluded that the use of OpenFL secured the collection of datasets from multiple sources. The diversity of the dataset improved the accuracy and reduced the bias of the DL model.
              \item \textbf{Fed-BioMed:} is an open source Python initiative specializing in applying FL to real-world medical applications \cite{fedbio}. It facilitated ML model training on decentralized data without the need for data sharing. Fed-BioMed can be used in healthcare for various purposes, such as dimensionality reduction in multicenter structural brain imaging data from different geographical locations \cite{silva2020fed,khan2023federated} and analysis of MRI data for prostate cancer diagnosis \cite{jablecki2021federated}. Cremonesi et al. \cite{cremonesi2023fed} presented results from implementing and training a prostate segmentation model with FL-Biomed, using data from three different medical institutions. Fed-BioMed also addressed many statistical and systems challenges in the field of biomedical research, demonstrating its capabilities for real-world applications of FL in hospital networks.
        \end{enumerate}
\subsubsection{Security and privacy (S)}
    Security and privacy are crucial in FL because the integrity of the model and the privacy of participants can be attacked by various types of attacks. Below are some of the security and privacy challenges and suggested methods to mitigate these attacks in FL:
        \begin{enumerate}[itemindent=20pt,labelsep=4pt,labelwidth=1em,leftmargin=2pt,label= \textbf{S\arabic*.}]
              \item \textbf{Privacy attacks: }While FL represents an advance toward collaborative learning with privacy, significant research challenges remain. One of the biggest privacy challenges is Gradient Inversion \cite{hatamizadeh2023gradient}. While FL avoids sharing private raw data, the model updates exchanged during training could potentially reveal sensitive information about the training data. Another challenge is the reconstruction model attack. An example of this attack is the Generative Adversarial Network (GAN). GAN-based attacks are a type of reconstruction attack that aims to reproduce private training data using GAN. The latter are trained using model updates or gradients from victim patients as feedback to refine the artificially generated data \cite{nair2023robust}. Techniques such as secure multiparty computation \cite{zhao2019secure} and homomorphic encryption \cite{naehrig2011can} can help mitigate these privacy attacks, but they have limitations. Protecting privacy while maintaining the usefulness and security of the model remains an ongoing challenge. 
              \item \textbf{Federated model attacks: }As FL is increasingly used in medical departments such as healthcare and IoT, securing these systems is critical. Federated models are subject to data poisoning attacks, where malicious participants send manipulated training data to degrade model performance. Ongoing research has been conducted to improve detection capabilities, for example by analyzing model performance changes or client update patterns to identify anomalies. However, new attack strategies are constantly emerging. This section provides an overview of these attacks: 
                \begin{itemize}
                      \item \textbf{Poisoning Sample:} refers to an attack in which malicious participants inject manipulated or biased data into the training process to degrade the performance of the global model. There are three categories of data poisoning attacks: targeted label flipping, random label flipping, and random input data poisoning in an FL environment  \cite{gupta2023novel}. Many researches proposed data poisoning attacks, for example Kasyap et al.  \cite{kasyap2023beyond} presented an inverted loss function and anti-training to maximally distort the training data. This technique resulted in the most effective data poisoning attack compared to other poisoning methods. Yang et al.  \cite{yang2023model} proposed a highly stealthy data poisoning attack called model shuffle attack (MSA), which is an attack that exploits the shuffling/scaling of model parameters to degrade the performance of FL systems. The attack remains undetected by bypassing common defenses. To resolve and detect intoxication attacks, many researches have been conducted, for example by Yang et al.  \cite{yang2023demac} introduced DeMAC, a novel defense method for FL against model poisoning attacks. By examining gradient norms of client updates, the authors found that DeMAC outperformed existing defense methods based on robust aggregation rules or anomaly detection in mitigating poisoning attacks. Tolpegin et al.  \cite{tolpegin2020data} proposed a defense method based on dimensionality reduction.  The idea is to extract relevant parameters from high-dimensional updates and use PCA to reduce the dimensionality and visualize the clusters to identify potential poisoning attacks. 
                      
                      \item \textbf{Label flipping:}  is an adversarial attack technique described in  \cite{sanchez2023robust} that intentionally changes the labels of some training data samples to incorrect values. This is an easy way for attackers to manipulate their local training data and corrupt the global model in FL. The authors simply swap the labels of some examples in their local data from a selected source class to a selected target class, while leaving the input functions unchanged  \cite{jebreel2022defending}. A slight flip of the label can be very damaging. Flipping less than 10\% of the labels can drastically affect model accuracy. Several studies are being conducted to detect label flipping using methods and techniques such as MCDFL  \cite{jiang2023data} and KPCA and K-Means Clustering \cite{li2021detection}. These techniques have shown promising performance in detecting this type of attack. In summary, label flipping is used to simulate malicious training data poisoning in FL and analyze the robustness of various aggregation algorithms against this attack. 
                      
                      \item \textbf{Backdoor attack: }This is an attack that allows attackers to hijack FL models for malicious purposes through targeted manipulation during training. They take advantage of the distributed nature of FL, where models from many participants are combined into a global model. An attacker uses some participant models to inject the backdoor \cite{bagdasaryan2020backdoor}. Many techniques and methods have been developed to defend backdoor attacks. For example,  Zhu et al. \cite{zhu2023adfl} proposed a defense method called adversarial distillation-based backdoor defense (ADFL) that uses GAN-based data augmentation. This method has effectively defended FL against backdoor attacks. Wang et al. \cite{wang2023scfl} proposed a method called SCFL. This method works by intelligently selecting the harmless gradients and excluding backdoor gradients, by using singular value decomposition (SVD) and K-Means clustering. Also, Cosine similarity is used to select the cluster that is most likely to contain only harmless gradients for aggregation. Sun et al. \cite{sun2019can} conducted several experiments on EMNIST dataset, and found that the defensive  norm clipping/thresholding had the best results against backdoor attacks.
                \end{itemize}
        \end{enumerate}

\subsubsection{Types of cancer (T) }
    Cancer is a group of more than 100 different diseases that can occur almost anywhere in the body. Doctors divide cancer into different types depending on where it starts. Here are some common cancers:
        \begin{enumerate}[itemindent=20pt,labelsep=4pt,labelwidth=1em,leftmargin=2pt,leftmargin=5pt,label= \textbf{T\arabic*.}]
              \item \textbf{Skin Cancer:} is a pathological condition characterized by the uncontrollable proliferation of abnormal skin cells. This disorder is caused by unrepaired deoxyribonucleic acid (DNA) in skin cells \cite{dildar2021skin} caused by exposure to ultraviolet radiation (UVR). It was the most commonly identified cancer in the United States \cite{chang2023nanoparticles}. There are several variants of skin cancer, including melanoma, squamous cell carcinoma, basal cell carcinoma and actinic keratoses \cite{ferguson2023risk}. 
              
              \qquad Several studies have shown that FL has improved skin cancer detection in various ways. Hashmani el al \cite{hashmani2021adaptive} have developed a flexible federated ML-oriented model for skin disease detection, which can be of great help to dermatologists in the preliminary diagnosis of skin malignancies and assessing their severity. Cai et al. \cite{cai2021many} proposed a federated deep skin CD model (FDSCDM) that used FL and DualGAN to tackle the challenges of data security and privacy concerns in medical IoT settings. The proposed framework demonstrated exceptional precision and area under the curve (AUC) in detecting skin cancer, while ensuring confidentiality and data privacy.

              \qquad Numerous research articles have used the practice of TL in skin CD. Hosny et al. \cite{kassem2020skin} used TL and image augmentation to propose a deep CNN to overcome the challenge of requiring a large number of labeled images for training. They showed that this technique outperforms other models in terms of accuracy, sensitivity, specificity and precision. Kondaveeti et al. \cite{kondaveeti2020skin} presented an analysis of the contribution of TL to improving the precision of skin lesion image classification in the HAM10000 dataset diagnosis using pre-trained CNNs as feature extractors. Ali et al. \cite{ali2021enhanced} used a pre-existing DenseNet model and fine-tuning it with their own skin cancer dataset. The authors also used a class-weighted and focal loss function to address the problem of class imbalance in the dataset.
              
              \item \textbf{Breast Cancer: }is the most commonly diagnosed cancer and the leading cause of cancer death in women. It is characterized by the uncontrollable growth of abnormal cells in the breast tissue \cite{dumalaon2014causes}.

              \qquad FL has been used in several researches, Roth et al. \cite{roth2020federated} conducted a study to develop a FL model for breast classification of mammography data from seven clinical institutions across the world. This study showed that the results of this model improved by 6.3\% compared to models trained on a central server, despite dissimilarities and differences between datasets across all sites. Jimènez-Sànchez et al. \cite{jimenez2023memory} proposed an approach: Fed-Align-CL, a decentralized model based on federated architecture. The proposed Fed-Align-CL model achieved the highest AUC and PR-AUC. Kumbhare et al. \cite{kumbhare2023federated} proposed a FL-based E-RNN model that collected mammography images of breast cancer and used the Densenet model for feature extraction. These extracted features are then classified using E-RNN for breast cancer detection.

              \qquad TL has been used in breast CD, Kumari et al. \cite{kumari2023magnification} used a pre-trained deep CNN architecture to perform breast cancer classification on two datasets. Densenet-201 architecture achieved the best results with classification accuracy of 99.50\% and 99.12\% from two datasets. Khan et al. \cite{khan2019novel} proposed a DL framework for the detection and classification of breast cancer using TL and fine-tuning. Classification was based on two separate sets of microscopic breast images. The authors combined three CNN architectures (GoogLeNet, VGGNet, and ResNet). The images in both datasets were captured by a microscope at different magnifications (100X, 140X, 200X, and 500X). The authors compared this framework with other CNN models and the proposed framework achieved the best results. 
                
              \item \textbf{Colon Cancer: }Commonly referred to as colon cancer, it is a malignant neoplasm that targets the colon or rectum. This disease is one of the most serious and deadly cancers in the world and is among the three most common worldwide \cite{pacal2020comprehensive}. In the Netherlands, colorectal cancer accounts for almost 10\% of all cases, with colorectal cancer being the second most common cause of cancer-related death \cite{ahmed2020colon}. 

              \qquad TL has been used in many researches to detect colon cancer. Murugesan et al. \cite{murugesan2023colon} proposed the use of YOLOv3 MSF as a TL approach to identify and annotate different stages of colon cancer. The authors used the VColonDB dataset in their study. Mehmood et al. \cite{mehmood2022malignancy} used a dataset of 25,000 histopathological images of lung and colon tissue and trained an AlexNet model, using 80\% of the images in each class for training purposes and 20\% for testing purposes. The proposed class selective image processing (CSIP) approach improved the accuracy of the model from 89\% to 98.8\%. DenseNet and SENet architectures were used by Gessert et al. \cite{gessert2019deep} to train pre-trained models to classify healthy colon and peritoneum tissue from confocal laser microscopy (CLM) images with AUC greater than 90\%. This research showed that TL achieved great results and can be used with other approaches to improve the accuracy of the model.

              \item \textbf{Brain Cancer: }refers to malignant and benign tumors that arise in the brain or surrounding tissues. The development of brain tumor can originate either from primary brain cells or from the metastasis of cancer cells from other regions of the body \cite{hormuth2022opportunities}. Glioblastoma multiforme (GBM), classified as grade 4, is the most dangerous and widespread brain tumor. Despite advances in therapy, the survival rate of GBM patients has remained low in recent decades, with an average survival time of less than two years \cite{al2022inflammation}. 

              \qquad In \cite{yi2020net}, Yi et al. proposed an efficient FL model SU-Net for brain tumor segmentation based on the encoder-decoder model. The approach was to collect brain MRI images from five medical institutions. The results of the study showed that the SU-Net model has the best performance, outperforming the baseline models (including FCN8s, standard U-Net and DeepLabv3+) in both AUC and DSC metrics across all datasets from five different institutions. Rehman et al. \cite{rehman2020deep}  trained various CNN models such as AlexNet, GoogLeNet and VGGNet on the Figshare dataset (MRI images). The authors reported that the VGG16 approach achieved the best results with an accuracy of 98.69\%. Talukder et al. \cite{talukder2023efficient} have trained several TL algorithms, including Xception, ResNet-50V2, InceptionResNetV2, and DenseNet201 on Figshare brain tumor dataset  and achieved accuracy scores of 99.40\%, 99.68\%, 99.36\%, and 98.72\% for each algorithm, respectively. Lyu \cite{lyu2022transformer} proposed the combination of convolutional layers and transformers for brain metastasis segmentation, the authors trained their model on MRI scans at Wake Forest School, the proposed approach also showed better performance than classical results in classification of brain metastases. Anaya-Isaza et al. \cite{jacob2023brain} applied multiple CNN models to BraTS 2018 and TCGA-LGG brain tumor detection datasets and concluded that DenseNet121, InceptionV3 and VGG19 produced the best results.
              
              \item \textbf{Lung Cancer:} is characterized as a neoplasm of the respiratory organs and is the leading cause of global cancer incidence and mortality,  with approximate of 2 million diagnoses and 1.8 million demises \cite{thandra2021epidemiology}. In spite of other possible factors such as air pollution, genetic predisposition and exposure to radon. Smoking has been identified as a key risk factor for lung cancer \cite{rudin2021small}.
              
              \qquad Heidari et al. \cite{heidari2023new} proposed a framework named FBCLC-Rad that integrated edge and blockchain for lung CD, this model was based on horizontal FL approach. The authors trained their proposed model on four distinct datasets: cancer imaging archive (CIA), Kaggle data science bowl (KDSB), LUNA 16, and a local dataset. The model achieved an overall accuracy of 98.9\%, a sensitivity of 98.7\%, and a specificity of 99.1\%. Liu et al. \cite{liu2023federated} proposed a FL approach for lung nodule detection using a 3D ResNet18 dual path Faster R-CNN network. The authors trained their proposed model on Luna16 dataset and achieved F1 score of 83.401\%. Ayekai et al. \cite{ayekai2022federated} trained a FL model with FedAvg strategy on Colon Cancer Histopathological Image Dataset, the authors used the CNN DenseNet121 model to train the data and achieved an accuracy of 94.5\% on the test set. The centralized learning method had an accuracy of 95.2\%. With these results, the authors concluded that the two approaches achieved very similar results despite protecting the privacy of the data.           
              
              \qquad TL has been used in several research articles on lung cancer. Fang et al. \cite{fang2018novel} proposed an approach in which a GoogLeNet-based CNN was trained using median intensity projections (MIPs). The CNN performed the classification with an accuracy rate of 81\% in identifying malignant and benign pulmonary nodules. Sajja et al. \cite{sajja2019lung} also proposed a pre-trained CNN based on GoogleNet to be used as a feature extractor. The model was trained on the lung image database consortium (LIDC) dataset with a test accuracy of 99.00\%.              
        \end{enumerate}

\subsection{Public Datasets}
\begin{figure*}[h]
    \begin{subfigure}{0.5\textwidth}
      \centering
      \includegraphics[width=.8\linewidth]{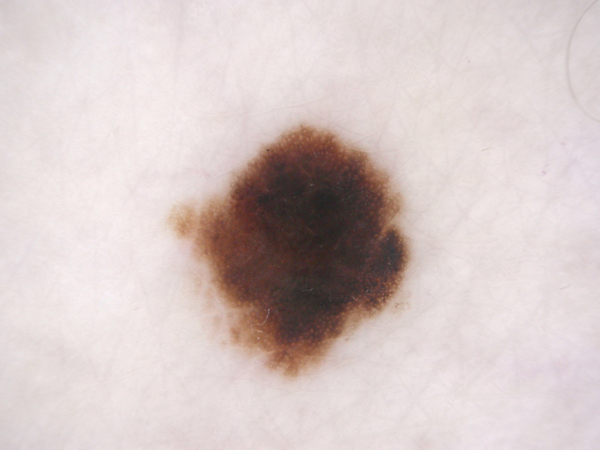}
      \caption{Melanoma skin tumor} 
    \end{subfigure}%
    \begin{subfigure}{0.5\textwidth}
      \centering
      \includegraphics[width=.8\linewidth]{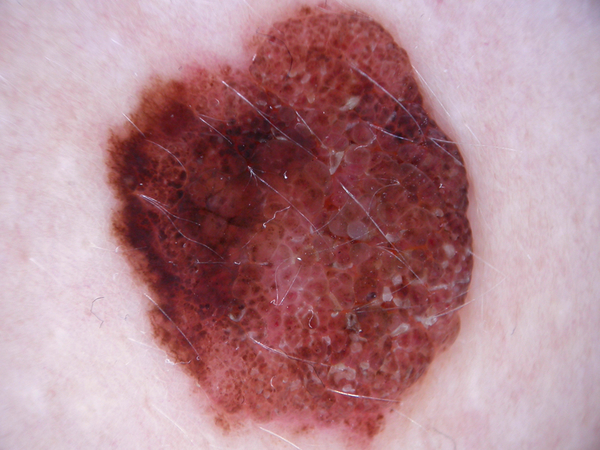}
      \caption{Melanocytic Nevi skin tumor}
    \end{subfigure}
    \caption{Sample of Skin Cancer MNIST: HAM10000 dataset \cite{codella2019skin,tschandl2018ham10000}}
    \label{fig:examdataset}
\end{figure*}

Publicly available datasets provide researchers with the opportunity to develop FL and TL learning techniques for CD based on medical image analysis. Table \ref{tab:datasets} contains a list of publicly available datasets covering various cancer types, including breast, brain, lung, skin and colon cancer. These datasets contain a range of medical images such as mammograms, MRI scans, CT scans and microscopic images with labels and annotations. Researchers can use these public datasets to develop and evaluate FL and TL approaches for CD tasks. Overall, these publicly available datasets are a critical resource for important advances in cancer research and improving CD. Fig. \ref{fig:examdataset} represents a sample from the skin cancer dataset HAM10000.
\subsection{Considerations for choosing between FL and TL}
There are several key factors to consider when deciding on the appropriate approach to build a strong FL and TL model:
\begin{itemize}
    \item  
FL enables model training on decentralized data residing on different devices without the need for direct data exchange. This improves data protection and reduces security risks. TL requires the aggregation of data on a central server, which raises greater privacy concerns.

\item 
FL is well suited for heterogeneous decentralized data. TL works better when the data is similar across all clients.

\item 
FL performance depends on the quantity and quality of local data. TL allows reusing a pre-trained model on a very large dataset, enabling better performance even on small local datasets.

\item 
With FL, higher communication costs are incurred between the central server and local devices during training. TL only requires a one-time transfer of the pre-trained model.

\item  
FL allows models to be customized to any local dataset. TL creates a more general model.
\end{itemize}

\subsection{Advantages and disadvantages of each method}

This section provides an overview of the advantages and limitations of FL and TL for CD based on image analysis. An analysis for each technique is used for each learning type. 

FL had several advantages for CD, including improving the diversity of the model and allowing personalized learning on client devices when using the FedAvg algorithm. FL also used fragmented medical data across sites while protecting privacy. As shown in Table \ref{tab:advatange}, using a 3D ResNet18 model trained with FedAvg. Bayesian modeling within FL handles recommendations for unseen groups and prevents collapsed representations. However, FL faces some limitations - the FedAvg algorithm can struggle to handle heterogeneous data distributions across clients. Small dataset sizes, as in an LSTM and Transformer FL model, limit performance. And convergence analysis remains challenging with algorithms like FedAvg.
\begin{table*}[t]
\centering
\caption{Public Cancer Datasets}
\renewcommand{\arraystretch}{1.5}
\label{tab:datasets}
\begin{tabular}{cllll}
\hline
URL & Name & Cancer Type & Description & \# of Samples \\
\hline
\href{https://www.kaggle.com/kmader/skin-cancer-mnist-ham10000}{D1} & HAM10000& Skin & \makecell[l]{A collection of multi-source  dermatoscopic images \\ of common pigmented skin lesions} & 10,015 images \\

\href{https://www.ebi.ac.uk/ega/studies/EGAS00000000083}{D2} & METABRIC & Breast  & \makecell[l]{A dataset of genomic and  clinical data from \\  breast cancer patients} & Over 2,000 patients \\

\href{https://www.bcsc-research.org/data/data-access}{D3} & BCSC & Breast & \makecell[l]{A dataset of mammography and  clinical data} & Over 2 million women \\

\href{http://www.cgga.org.cn/download.jsp}{D4} & CGGA & Brain & \makecell[l]{A dataset of genomic and clinical data \\ from glioma patients} & Over 1,000 patients \\

\href{https://www.med.upenn.edu/cbica/brats2021/data.html}{D5} & BraTS & Brain & \makecell[l]{A dataset of MRI scans of brain tumors for the \\ purpose of developing algorithms \\ for automated segmentation of brain tumors} & 1,000+ patients \\

\href{https://portal.gdc.cancer.gov/projects/TCGA-COAD}{D6} & TCGA & Colon & \makecell[l]{A dataset of genomic, transcriptomic, and epigenomic \\ data from colon cancer patients} & Over 400 patients \\

\href{https://www.ncbi.nlm.nih.gov/geo/query/acc.cgi?acc=GSE17538}{D7} & GEO & Colon  & \makecell[l]{A dataset of gene expression  data from \\  colon cancer patients} & Over 200 patients \\

\href{https://portal.gdc.cancer.gov/projects/TCGA-LUAD}{D8} & GDC& Lung & \makecell[l]{A dataset of genomic and  clinical data \\ from over 1,000 lung cancer patients} & Over 1,000 patients \\

\href{https://dcc.icgc.org/releases/PCAWG}{D9} & ICGC & Lung & \makecell[l]{A dataset of genomic and  transcriptomic data \\ from lung cancer patients} & Over 500 patients \\

\href{https://www.kaggle.com/nodoubttome/skin-cancer-isic}{D10} &  ISIC & Skin & \makecell[l]{A dataset of images of malignant and benign \\ oncological diseases} & 2,357 images \\
\hline
\end{tabular}
\end{table*}

\begin{table*}[t]
\caption{Advantage and limitations of FL and TL for CD}
\label{tab:advatange}
\renewcommand{\arraystretch}{2}
\begin{tabular}{lllll}

\hline
Ref & Type & Technique & Advantage & Limitation \\ \hline

\cite{liu2023federated} & FL & \makecell[l]{3D ResNet18 \\ + FedAvg} & \makecell[l]{Utilizes fragmented medical data, \\ protects privacy }& Effect of client number \\

\cite{fang2018novel} & TL &\makecell[l]{ GoogLeNet,\\ MIPs} & \makecell[l]{Improves accuracy and convergence \\rate compared  to training a CNN \\ from ground up} & \makecell[l]{Develop a fast, accurate and \\ stable DL system for \\ lung cancer detection from \\ CT scans} \\

\cite{cai2021many} & FTL & DGAN & \makecell[l]{Augments limited data, protects privacy }& \makecell[l]{ Impact of client number,\\  malicious clients} \\

\cite{li2023vision}& TL & ViT &  \makecell[l]{Captures global context, outperforms \\ CNN models} & \makecell[l]{ Requires large datasets  for \\ pretraining} \\ 

\cite{paya2022automatic} & TL & \makecell[l]{SCL} & \makecell[l]{Improves embryo viability and \\ quality classification} & Limited dataset from single clinic \\ 

\cite{wang2021survey} & TL & \makecell[l]{K-means, spectral \\ clustering} & \makecell[l]{Can produce good clustering results \\ when instance size is small by transferring \\ knowledge from source domain} &\makecell[l]{ Most research focuses on \\ classification and recognition,\\ not clustering} \\

\cite{das2021multi} & FL & \makecell[l]{TDCD}& \makecell[l]{Communication efficient, exploits vertical \\ and horizontal data partitions} & Analysis on non-IID data needed \\

\cite{ek2020evaluation} & FL & FedAvg & \makecell[l]{Improves model diversity and provides\\  personalized learning on client devices} &\makecell[l]{  Difficult to handle data \\  distributions across clients} \\

\cite{sharma2023deep} & TL & VGG-16 & \makecell[l]{Achieves high accuracy for pneumonia \\ detection  from chest X-rays} & \makecell[l]{ Limited comparison to other \\ best models }\\

\cite{ikechukwu2021resnet} & TL & ResNet-50, VGG-19 & \makecell[l]{Improved performance over training \\ from ground up} & Small dataset size \\ 

\cite{panthakkan2022concatenated} & TL & \makecell[l]{Xception and \\ ResNet50} &  \makecell[l]{Achieves high accuracy for skin \\ cancer classification} & Image dataset is imbalanced \\

\cite{zidan2023swincup} & TL & \makecell[l]{Swin Transformer \\ and CNN} & \makecell[l]{Accurate segmentation, captures \\ global context} & Dataset size \\ 

\cite{raval2023comprehensive} & TL & \makecell[l]{AlexNet, VGGNet,\\ Inception} & \makecell[l]{DenseNet achieved best performance \\ for skin CD} & \makecell[l]{Limited dataset available  for \\ oral cancer, making \\ detection difficult} \\

\cite{dey2022screening} & TL & DenseNet121 & Accurate classification & Single dataset \\

\cite{hanser2023federated} & FL & \makecell[l]{Model-driven and \\ data-driven FL} &  \makecell[l]{Improves model performance and \\applicability domain} & \makecell[l]{ Communication efficiency \\ challenges} \\

\cite{su2021hierarchical} & FL & SCGDA & Utilizes sample and feature diversity & Convergence analysis needed \\
\cite{lee2022bayesian} & FL &Bayesian modeling & \makecell[l]{Handles recommendations for unseen groups.\\ Better prevents collapsed representations} & \makecell[l]{Slower convergence compared \\ to some existing models} \\
\cite{kamei2023comparison} & FL &\makecell[l]{ LSTM, Transformer \\ and FedAvg} & \makecell[l]{Decentralized approaches address data privacy \\ and security issues} & Small dataset size \\ 
\cite{sun2022decentralized} & FL & FedAvg & \makecell[l]{blackuces communication bottlenecks, more \\ robust  to node failures and privacy attacks} & Convergence analysis is challenging \\ \hline
\end{tabular}
\end{table*}

TL also provides advantages for CD. TL improves accuracy and convergence rate compared to training a model from ground up, as demonstrated through a GoogLeNet model with Median Intensity Projections (MIP) for lung cancer. TL used features learned on large datasets by models like VGG16 and ResNet50 to improve performance on new tasks. ViTs like Swin Transformer and ViT capture global context from medical images to aid analysis. However, TL has some limitations. ViTs require large datasets for pretraining to be effective. Models like DenseNet121 generalize poorly to new patients or classes. And individual institutions often have limited datasets, as seen in a supervised contrastive learning model for embryo images. Table \ref{tab:advatange} highlights the main advantages and limitations of FL and TL.

\section{Challenges and Open Issues}
Despite TL comes with a set of advantages in the medical department, using TL for CD can present several challenges and limitations.

\color{black}
\subsection{Domain shift}

Data distribution within the CD target domain may show notable inconstancy when compared to the pre-trained model's source domain. This can result in a domain shift phenomenon, which has the potential to negatively impact the model's overall performance \cite{gu2019progressive,stacke2020measuring}. Not having enough data from the target domain can cause TL and FL models to perform poorly.
 
Domain shift is caused because of variations in staining methods, differences in scanning devices between datasets, number of classes, variation in object orientation (differences in image content, view angle, brightness, noise, color etc...) and number of channels in datasets \cite{zoetmulder2022domain}. Even with the same source and target domains, Zoetmulder et al. \cite{zoetmulder2022domain} findings on the impact of domain shift on lesion detection accuracy were inconsistent. This implies that there is still more to learn about the effects of domain shift. Additionally, TL models often experience a drop in performance when tested on data from a different distribution than the training data. However,  the authors in \cite{fogelberg2023domain}  tried to solve domain shift by grouping images from public dermoscopic skin cancer datasets into different domains to create domain-shifted test sets. Another way to overcome domain shift is using extensive augmentations like PatchShuffling to improve the learned representation and fine-tuning pre-trained model on source dataset can handle domain shift \cite{vuong2022impash}.
 
In summary, TL and FL face serious challenges under domain shift, which must be approached with caution due to the variability in data distributions, feature spaces, tasks, and data access.

\subsection{Computational and Hardware Limitations}
Healthcare organizations may face limitations in their hardware capabilities and there is a need for constant monitoring and updating to maintain the performance of the FL and TL models. For example, hardware limitations can have a negative impact on TL. This could negatively impact the data collection process and the use of TL techniques. For example, in \cite{marathe2017performance,whatmough2019fixynn}, the authors emphasized that hardware limitations such as limited memory or limited computing capacity can make training complex DL models difficult and it is a challenging process. This could limit the type and size of neural network models that can be effectively trained.

In FL, devices with slower processing speeds can cause delays in the training process as they do not complete their assigned training on time. This delay can negatively impact the overall progress of model training. Additionally, devices with limited memory and bandwidth may struggle to run complex FL algorithms efficiently, potentially resulting in suboptimal model performance \cite{abreha2022federated,imteaj2021survey}. In addition, the heterogeneity of computing, storage and communication capabilities of different devices can cause latency issues in on-device distributed training \cite{imteaj2021survey}. Energy is then used by the devices for local model training and communication. In general, less energy efficient devices and radios have lower processing power and bandwidth. This causes batteries to drain quickly when using wireless devices \cite{tran2019federated}. Therefore, implementing FL systems often requires overcoming complex system challenges such as unreliable device connectivity, interrupted executions, and slow convergence rates compared to learning on centralized data \cite{wang2019adaptive,yang2020federated}.

In conclusion, limited hardware resources may slow down the training process significantly, making it challenging to test diverse techniques efficiently and develop powerful and accurate FL or TL models. 

\subsection{Limited availability of medical annotated data}


The acquisition of annotated medical data is an essential prerequisite for the effective training of ML models. Although, within the domain of CD, the attainment of annotated data can prove to be a difficult task due to ethical and privacy-related issues and the process of medical data annotation requires expertise \cite{shamshiri2023compatible}. This can subsequently constrain the quantity of data that is accessible for TL. This lack of data can cause domain shift and different distribution between training and test data that are independent of dataset size, in which can affect drastically the performance of TL \cite{alzubaidi2021novel}. Moreover, The limited number of annotated data and small dataset size can lead TL to problems such as miscalibration between classes and increased generalization error, especially for imbalanced classes \cite{abbas2020detrac}.

The field of healthcare data analysis faces several challenges related to data insufficiency, heterogeneity, and labeling.  For example, studies on colon cancer outcomes suffered from insufficiency of labels in datasets and heterogeneity of studies \cite{pacal2020comprehensive}. Also, the lack of large labeled cancer medical imaging datasets affected the training of FL and complicated the convergence during federated optimization \cite{rieke2020future}. Finally, handling non-IID data distributions and imbalanced datasets across clients is another challenge that needs to be addressed in healthcare data analysis \cite{rehman2023federated}. So for that, techniques such as data augmentation and collaborative efforts have been done to build annotated and standard datasets for CD \cite{rehman2023federated}. In general, limited data allows TL or FL models to overfitting to the specified features of the training set, non-IID data distributions and imbalanced datasets. Leading to poor general applicability and eventual model failure.

\subsection{Heterogeneity of cancer} Cancer is a very complex and diverse ailment characterized by numerous subcategories and developmental phases (malign and benign tumors...), which presents a challenge to TL and FL capacity to generalize across dissimilar cancer subtypes and stages. In a study by Jaber et al. \cite{jaber2020deep} An image-based DL intrinsic molecular subtype classifier was developed to classify breast tumors into different subtypes based on H\&E-stained biopsy tissue sections. The classifier was able to correctly subtype the majority of samples in a retained set of tumors, but in many cases significant heterogeneity in the assigned subtypes was observed across patches within a single overall slide image. This heterogeneity impacted the accuracy of the classifier and may require more detailed subtype analysis to improve classification results.  Nyman et al. \cite{nyman2023spatially} highlighted that heterogeneity within tumor samples (microheterogeneity) is common and can impact CD using DL. Quantifying and accounting for this heterogeneity is important for improving cancer classification, prediction, and treatment planning. Heterogeneity within a tumor can lead to misclassifications or lower accuracy for DL models \cite{inglese2017deep}.
\subsection{Data Privacy and Security}
FL allows different organizations to work together and train DL models without sharing their sensitive data. Unfortunately, there are several risks associated with this approach, such as model poisoning, label flipping, and backdoor attacks \cite{yang2023demac,sanchez2023robust,bagdasaryan2020backdoor}. These threats can compromise the integrity and accuracy of the models being developed. To avoid these risks, it is important to detect any malicious activity early on and take appropriate action. Moreover, FL also presents some unique challenges when it comes to data privacy, security, and dealing with limited datasets. These issues can be tackled by safeguarding confidential information and minimizing communication expenses \cite{khan2021federated}. However, strict privacy policies and restricted data access pose limitations for collaborative model building across institutions. Implementing FL can be a complex process due to high communication overhead and intricate system requirements. Therefore, addressing these obstacles is critical to ensuring that FL succeeds in practical settings and in a healthy environment.

\subsection{Interpretability and Algorithm Challenges}

Adapting powerful TL models like transformers and CNN with FL to computer vision can improve the accuracy of image analysis tasks and the performance of decentralized learning algorithms, but big issue remains: Interpretability.

The decentralized design of FL creates difficulties for traditional model interpretation approaches since they usually need complete access to all features, training data, and internal model components \cite{wang2019interpret}. It is difficult to interpret why a single global model works well or poorly for different clients, without interpretability in the client selection and aggregation process, it is hard to understand and improve the FL system \cite{qin2023reliable}. In TL and despite it is difficult to understand how the TL models make decisions. Interpretability helps understand the interface between source and target domains. For this reason, the authors in \cite{kim2019structure} proposed a feature network (FN) approach that uses interpretable features defined by humans to improve interpretability in TL and improve the results. This address challenges in TL like negative transfer, insufficient target data, and lack of transparency \cite{mao2022interpretable}. As a result, new strategies must be created to offer transparent and interpretable explanations about how FL and TL models work while still maintaining the privacy of the data and improve the accuracy.

\subsection{3D Imaging}
Medical image analysis can be challenging, especially when it comes to 3D images. One of the most important challenges is data availability, this is due to the difficulty and expense of obtaining 3D images compared to 2D images, which prevents the availability of big and diverse datasets for training TL models \cite{chen2019med3d}. Despite techniques such as data augmentation are used to add number of samples but the authors in \cite{gupta2020performance} stated that 2D data augmentation techniques like rotations, flips and crops don't directly translate to 3D. Furthermore, in comparison to 2D images, 3D images are more complex and require more advanced techniques to analyze \cite{cheplygina2019not}. Another issue is data heterogeneity, this is due to diverse modalities, dimensionalities, and features that can be found in 3D images, along with differences in acquisition and demographics \cite{wang2016new}, which can affect negatively TL. For example, the authors in \cite{zhou2019models} stated that 3D medical images can be highly complex, analyzing 3D structure adds complexity compared to 2D images. 

Even with these challenges, researches  are actively investigating to improve 3D medical image analysis through new approaches and techniques that could have a significant positive impact on TL training.

\section{Case Studies}
 FL and TL have shown promising applications in CD based on image analysis. The following subsections provide more details on current applications, future directions, and limitations of these methods:

\subsection{Example of FL in CD}
\begin{figure*}[t!]
\begin{subfigure}{0.5\textwidth}
  \centering
  \includegraphics[width=1\linewidth]{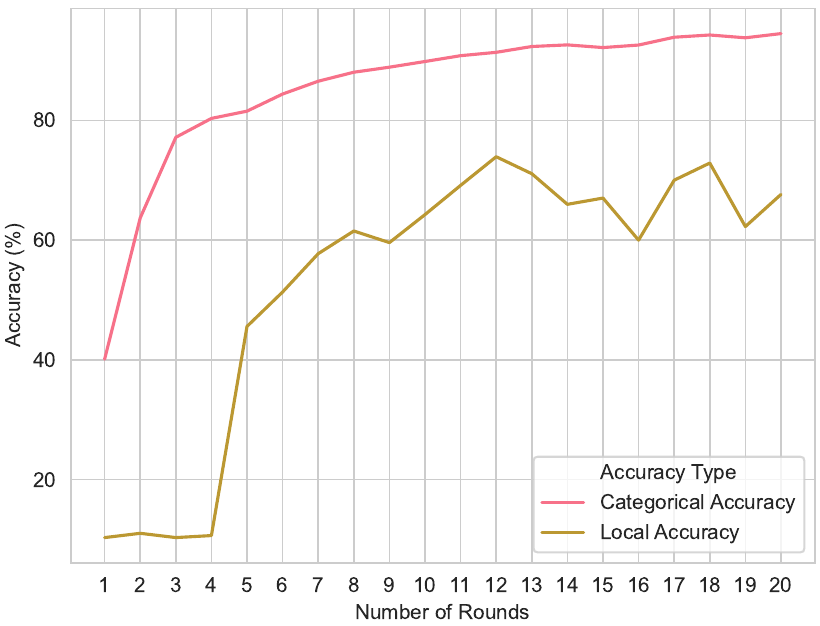}
  \caption{Accuracy} 
\end{subfigure}%
\begin{subfigure}{0.5\textwidth}
  \centering
  \includegraphics[width=1\linewidth]{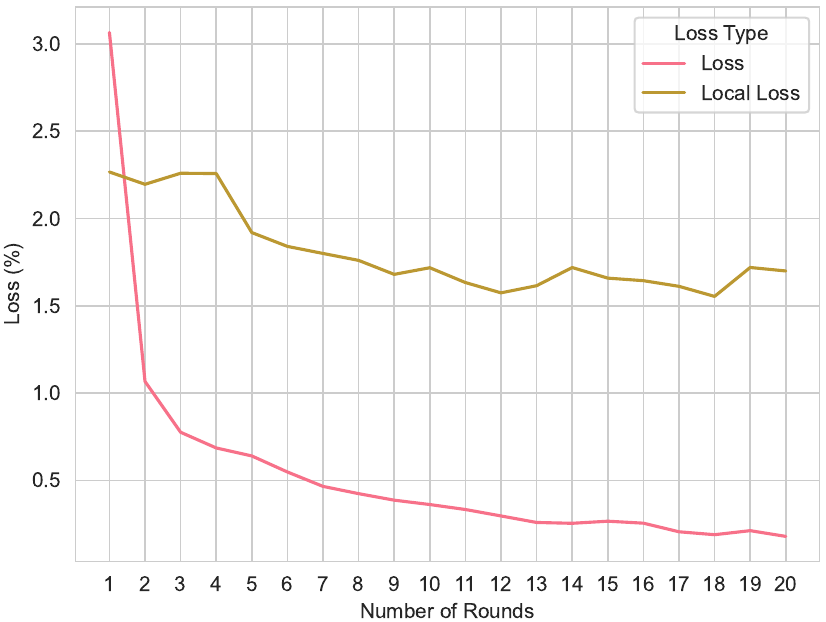}
  \caption{Loss}
\end{subfigure}
\caption{Performance (Accuracy and Loss) of FL based model with ResNeXt50 for detecting colorectal cancer in histopathological images \cite{arthi2022decentralized}}
\label{fig:accuracyexmp}
\end{figure*}

Kareem et al. \cite{kareem2023federated} proposed a framework for medical image detection using FL where training data was divided among virtual devices representing medical institutions. Pre-trained models like AlexNet, DenseNet, ResNet-50, Inception, and VGG19 were used in the FL framework. The FL ResNet-50 model achieved 93\% accuracy, while VGG19 achieved 97.94\% accuracy for pneumonia detection when combined with an alternative classifier. FL integrated with TL models showed promise for enhancing disease detection in medical images while ensuring data privacy. However, more details on the data sources used for training would help better assess the proposed model's performance. Repetto et al. \cite{repetto2022breast} proposed an FL model using goal programming to optimize model performance across nodes for breast CD. Experiments were conducted on the Wisconsin Breast Cancer Dataset. Compared to the initial model, the FL model showed significant 15-27\% improvements on metrics like J-index and AUC when evaluated on test data. The federated model also outperformed models trained at individual sites. The results demonstrated the efficacy of the proposed FL approach for breast CD. Pati et al. \cite{pati2022federated} applied FL to detect tumor boundaries in glioblastoma, a rare brain cancer, using 1963 MRI scans from 71 global sites. Model performance was evaluated on test data from participating sites and completely unseen data. The FL model showed 15-33\% improvements in Dice scores over the initial model, demonstrating its efficacy. However, only a single 3D-ResUNet architecture was tried. Evaluating other neural network architectures could have found an even better model for this use case. Arthi et al. \cite{arthi2022decentralized} proposed a FL-based model for detecting colorectal cancer in histopathological images while ensuring privacy. The authors used  heterogeneous datasets from multiple healthcare facilities with 0.1 million sample histological images of human colon cancer and healthy tissue with 9 classes. The authors applied various CNN models and found that ResNeXt50 achieved the highest accuracy of 99.53\%. They then implemented ResNeXt50 on FL, resulting in an accuracy of 96.045\% and an F1 score of 0.96. Fig. \ref{fig:accuracyexmp} representes the performance of the study across 20 communication rounds, with 10 epochs per round. The proposed workflow involved acquiring a dataset, image processing, and dividing the dataset into train, test, and validation data in which the dataset was distributed among local devices for training and testing to use the FL architecture. Twenty communication rounds between local and global models were performed to aggregate the trained models. We conducted experiments using Google Colaboratory \cite{bisong2019colab} notebooks with a Tesla K80 GPU backend accelerator and 12GB of RAM allocation. As shown, we were able to extract the accuracy and loss of ResNeXt50 on FL for detecting colorectal cancer in
histopathological images in Fig. \ref{fig:accuracyexmp}.

\subsection{Example of TL in CD}

\begin{figure*}[t!]
    \begin{center}
    \includegraphics[width=0.7\linewidth]{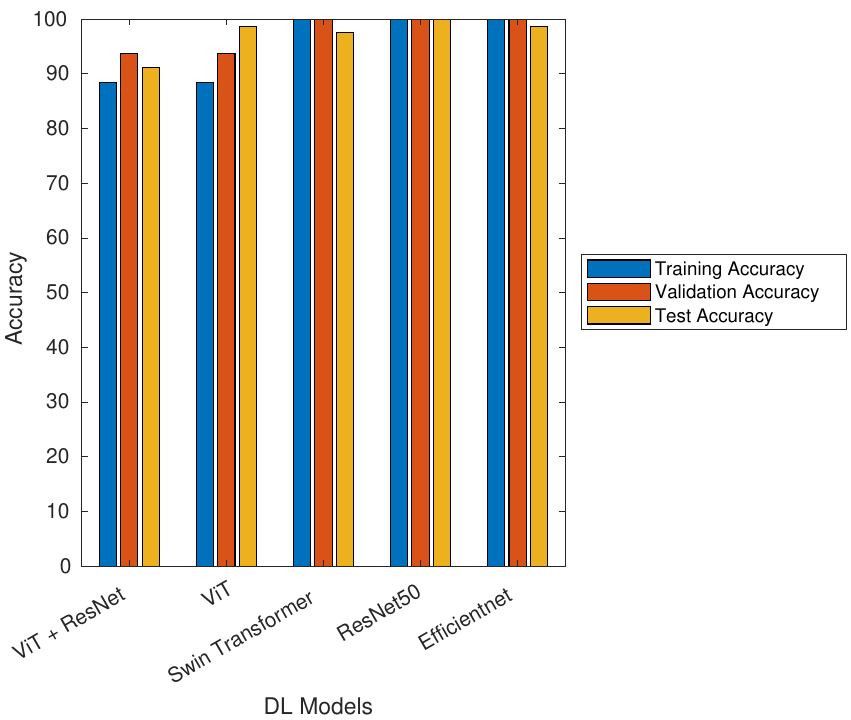}\\
    \end{center}
    \caption{SI-ViT and CNN models for Pancreatic Cancer MICCAI Image Classification based on accuracy \cite{zhang2022shuffle}.}
    \label{fig:compViT}
\end{figure*}

Kassani et al. \cite{kassani2022deep} conducted a comparison study of TL techniques for classifying colorectal cancer images. They used five pre-trained models (DenseNet121, DenseNet201, InceptionV3, MobileNetV2, ResNet50) with knowledge from ImageNet. Three TL techniques were compared: using the original network, fine-tuning, and adding convolutional layers. The TCGA CRC DX dataset was used. MobileNetV2 achieved the highest accuracy around 98\%. TL showed potential to improve cancer classification, though more techniques like using Imagenet and evaluation metrics could give a more comprehensive analysis. Luo et al. \cite{luo2023systematic} also studied different TL techniques like fine-tuning and adding convolutional layers for classifying colorectal cancer images. Cross-validation assessed performance. Adding convolutional layers outperformed other methods. TL was found efficient for small medical datasets and could reduce workload. More background on colorectal cancer's significance would have helped readers. Freitas et al. \cite{freitas2022detection} applied TL to pretrain models (AlexNet, VGG16, GoogLeNet, ResNet) for diagnosing bladder cancer in cystoscopy images. Capnets achieved 96.9\% accuracy, outperforming other TL models. TL increased performance over traditional techniques without requiring extensive data. A larger dataset and comparison to other DL models could have provided more insights. Aziz et al.  \cite{azizi2017transfer} investigated using TL for prostate CD from ultrasound data. TL could compensate for differences between data types. Cancerous regions were accurately identified, showing promise. But evaluation on other cancers would have been beneficial. Zhang et al. \cite{zhang2022shuffle} proposed a Shuffle Instances-based ViT (SI-ViT) approach that reduced image perturbations and improve cross-instance modeling. SI-ViT achieved significantly higher accuracy of 94\% compared to CNNs and basic ViT 91\% on a dataset of 1773 cancer and 3315 normal images. Fig. \ref{fig:compViT} demonstrates the results of our implementation of various models on MICCAI datasets using the code provided by Zhang et al. \cite{si2022milsicode}. The experiments were conducted using Google Colaboratory \cite{bisong2019colab} notebooks with a Tesla K80 GPU backend accelerator and 12GB of RAM allocation. As shown, we were able to replicate the findings that SI-ViT effectively extracted pathological patterns for ROSE diagnosis \cite{zhang2022shuffle}.

\section{Future Directions}

Testing FL and TL approaches in a real clinical setting is critical before deployment. It is also important to evaluate the performance, bias, interpretability and accountability of the model for implementing these techniques into clinical practice. Overall, FL and TL are exciting directions for the further development of privacy-preserving and precise AI in healthcare. These are future directions of this work:
\color{black}
\subsection{FL for non-IID and Heterogeneous Data:}

Dealing with non-IID and heterogeneous data is a critical challenge in FL, especially in the medical domain. Medical data is often distributed across federated clients, leading to non-uniform data distributions due to differences in data collection methods, user demographics, environmental conditions, and device configurations \cite{li2023heterogeneity}. This imbalanced distribution of data can create generalized models that are flexible to different use cases, but their performance is drastically impacted \cite{cheng2023gfl}. To address this issues, researchers have proposed various solutions. One promising approach is the Label-wise Clustering algorithm (FedLC) \cite{rehman2024fedcscd}, which guarantees convergence among local clients that hold unique data distributions, offering robust convergence on highly skewed and biased non-IID datasets \cite{rehman2024fedcscd}. Another study focuses on mitigating the negative impact of batch normalization (BN) on FL performance in the presence of non-IID data. The proposed FedTAN method aims to achieve robust FL performance under various data distributions \cite{wang2023batch}. Moreover,  Zhang et al. \cite{zhang2023tackling} proposed FedGH, an approach that  addresses local drifts caused by non-IID data and device heterogeneity by harmonizing gradients during server aggregation. FedGH has demonstrated impressive results on FL baselines across various benchmarks and non-IID scenarios. In addition to addressing non-IID data, researchers have explored data synthesis techniques to augment and improve FL performance. Li et al. \cite{li2024feature} proposed a novel data synthesis method called hard feature matching data synthesis (HFMDS), which optimizes synthetic data to be task-relevant and privacy-preserving by matching the class-relevant features of real data. When integrated with FL (HFMDS-FL), this method consistently outperformed baseline methods in terms of accuracy, privacy preservation, and computational costs.

\subsection{Large language model for FL:}
Large language models (LLMs) have emerged as a transformative technology in natural language processing, demonstrating remarkable capabilities in understanding and generating human-like text \cite{thirunavukarasu2023large}. The combination of LLMs and FL offers a promising future direction, sharing the powerful capabilities of LLMs while addressing the privacy concerns associated with their training \cite{bose2023fully}. Several studies have combined both techniques. For example, Sadot et al. \cite{sadot2023novel} have proposed a novel approach that combined FL with BERT, a large language model, and 1D-CNN for multi-label text classification on a large text dataset. Compared to centralized training on the entire dataset, the federated setup divided the data into groups and demonstrated lower computational power requirements while achieving higher accuracy, precision, F1 score, and lower Hamming loss for an equivalent global model. Kuang et al. \cite{kuang2023federatedscope}  introduced FederatedScope-LLM (FS-LLM), a comprehensive package for fine-tuning LLMs in FL settings. FS-LLM consists of three main components, LLM-BENCHMARKS; LLM-ALGZOO; LLM-TRAINER. The authors discussed challenges such as optimizing communication and computational resources, handling various data preparation tasks, and addressing distinct information protection demands. Liu et al. \cite{liu2023differentially} introduced DP-LoRA, that integrates differential privacy (to guarantee data privacy) and low-rank adaptation (to reduce communication overhead during distributed training). The experimental results demonstrated the effectiveness of DP-LoRA in maintaining strict privacy constraints while minimizing communication overhead. Ye et al. \cite{ye2024openfedllm} proposed OpenFedLLM combining instruction tuning, value alignment for LLMs, FL algorithms, and support for multiple datasets. Empirical studies using OpenFedLLM showed that models trained with FL consistently outperformed those trained individually on private data, achieving up to 12\% improvement on benchmarks like MT-Bench for general datasets. All these studies provides insights and suggests new directions for future work in the area of FL.

\subsection{Collaborative FL:}

Collaborative FL is emerging as a promising approach to enable secure multi-party collaboration while preserving data privacy in a decentralized, trustworthy, and flexible manner. This paradigm allows participants to collaboratively train a model without frequently sending local models or gradients to a central server, thus ensuring the privacy of their data. Recent advancements in this field have demonstrated the potential of Collaborative FL in various domains, particularly in healthcare applications \cite{abou2022collaborative}. Moving forward, Collaborative FL is poised to play a pivotal role in enabling collaborative training using synthetic data generated by GANs. Frameworks like FedCSCD-GAN  combined optimized FL and GAN-based optimization to address privacy, data quality, and model performance challenges \cite{li2023point}. By using synthetic data, such approaches can improve overall performance, stabilize classifiers, and prevent overfitting, while ensuring data privacy. In the case of cancer diagnosis and subtype prediction, Collaborative FL has shown promising results. Techniques like HistFL \cite{rehman2024fedcscd}, which incorporate attention-based multiple instance learning (MIL) and differential privacy, have demonstrated improved performance in collaboratively training models on multi-site whole slide images (WSIs). Additionally, collaborative healthcare diagnostic systems have the potential to improve local diagnosis performance, particularly for resource-constrained institutions with lower-quality imaging data \cite{gao2023federated}. Furthermore, Collaborative FL has been successfully applied to breast cancer detection, achieving impressive F1-scores of 97\% \cite{almufareh2023federated}, showcasing its advantage over traditional centralized approaches. As research in this area continues, it is anticipated that Collaborative FL will play a pivotal role in enabling secure and privacy-preserving collaboration across various domains, facilitating the development of robust and accurate models while protecting sensitive data.

\subsection{Decentralized Training Paradigm for FL:}

In decentralized FL, there is no central server that aggregates the model updates from the local devices. Instead, the devices communicate with each other and aggregate the updates in a peer-to-peer fashion. This approach can further enhance data privacy and security as there is no need to transmit any data, even in encrypted form, to a central server. 

Roy el al. \cite{roy2019braintorrent} proposed an approach called BrainTorrent, in which there is no central server that aggregates model updates from the different sites. Instead, sites share and aggregate model weights directly in a peer-to-peer fashion 
\color{black}

\color{black}
\newpage

without a central coordinator. The authors found that BrainTorrent showed good results and outperformed traditional FL with a global server. The study in  \cite{tedeschini2022decentralized} discussed the use of decentralized FL for healthcare networks, specifically in the context of brain tumor segmentation. The authors propose a real-time platform for integrating network and FL functions, and validate the proposed solution in a real-world deployment. In \cite{wicaksana2022customized} the authors proposed CusFL as a Personalized FL framework to handle the inter-client data heterogeneity issue in traditional FL, the authors evaluated CusFL on two multi-source medical image classification tasks: prostate cancer identification and skin lesion classification, demonstrating its superiority over traditional FL in handling the inter-client heterogeneity present in decentralized medical data. Decentralized FL can be particularly useful in scenarios where there is no trusted central authority or where the communication with a central server is not feasible due to network constraints.
\subsection{Domain Adaptation for TL:}
The main advantage of using domain adaptation is that it helps improve the general performance of TL models when there is a domain shift between the training and testing datasets \cite{han2023tl}. The domain adaptation aspect arises from the ability to effectively integrate and transfer knowledge from data-rich source domains to data-poor target domains, without negative transfer when the domains are not closely related \cite{hajiramezanali2018bayesian,choudhary2020advancing}. For example, Gu et al. \cite{gu2019progressive} explored extending the cycle consistent adversarial networks (CycleGAN) approach to handle larger structural changes between source and target domains, as the authors note that the model struggles with large differences. The authors applied the domain adaptation approaches to skin lesion classification where training data is limited. This has led to higher classification accuracy on the target dataset compared to just training on the original target data. You et al. \cite{you2019towards} applied  deep embedded validation (DEV) for model selection in domain adaptation tasks specific to cancer detection, where the source and target domains differ in imaging modalities, scanner types and patient populations. DEA improved model selection and domain adaptation performance for TL in CD tasks involving different imaging domains. Wang et al. \cite{wang2023iterative} proposed using a gaussian kernel-based distance constraint  (DDA) to  reduce the distance between the source domain and target domain in the feature space. This is a type of discrepancy-based domain adaptation. It helped decouple the motion-related features from user-specific features, this has improved TL. So domain adaptation techniques aims to reduce the cross-domain distribution shift by adapting source knowledge to the target domain, enabling transfer of learning across domains.

\subsection{Multimodal TL:}

Multimodal TL is an emerging paradigm that shows promise for various applications, particularly in the medical field. It refers to the process of transferring knowledge from one or more pre-trained models across different data modalities (such as text, images, audio, etc.) to a target model operating on multiple modalities simultaneously \cite{azher2023assessment}. The main advantage of multimodal TL is that it allows the target model to learn from different types of data more effectively, thus overcoming the challenge of limited data availability in certain modalities. In the field of medical imaging, multimodal TL has demonstrated its potential in various tasks such as cross-modal retrieval (CMR), prostate cancer detection and brain image analysis. For example, the deep multimodal TL (DMTL) approach   \cite{zhen2020deep} transferred knowledge from known categories to new categories by learning distinguishing features from both annotated labels and pseudo-labels, thereby enabling multiple baselines and existing methods with respect to the MAP exceeds scores for CMR. In addition, researchers have investigated the fusion of features from different input modalities, such as: B. different MRI modalities, to obtain a more comprehensive representation of the clinical condition of interest \cite{yuan2019prostate}. In the field of brain image analysis, multimodal deep TL techniques have been proposed to utilize different viewing planes (axial, sagittal and coronal) in MRI brain images \cite{azher2023assessment}. These techniques include novel multimodal feature encoders and adaptation techniques to address the distribution shift between training and test sets to improve overall performance. Furthermore, multimodal TL has played a crucial role in improving the segmentation accuracy of hypopharyngeal cancer risk areas in medical images \cite{zhang2023twist}. Approaches like Twist-Net combine low-level detail and high-level semantic information from different feature maps, resulting in improved segmentation performance. Looking forward, multimodal TL will play a crucial role in overcoming the challenges of limited data availability and improving the performance of various medical imaging tasks. By using complementary information from multiple data modalities such as X-ray, ultrasound and CT scans \cite{horry2020covid}, multimodal TL techniques can potentially open new avenues for more accurate and robust analysis and diagnosis of medical imaging.

\subsection{Attention Mechanism for TL:}
Future studies could investigate more sophisticated attention modules that can better capture the intricate patterns and relationships present in medical imaging data. For instance, self-attention mechanisms have been explored using transformer architectures or graph attention networks \cite{khan2023transformers}.
Furthermore, instead of simply freezing shallow layers during TL attention can be used to guide the layers or weights to transfer and fine-tune them based on the target task and dataset. This could lead to more efficient and effective transfer, and while transferring between related tasks such as different cancer types showed benefits, it would be valuable to study attention-based transfer between radically different imaging domains (e.g., transferring from natural images to medical images). This could open up new sources of pretrained models for use. 
Moreover, using TL on multiple complementary imaging modalities (for example, MRI, CT, and ultrasound) by fusing their representations could provide more comprehensive models \cite{yu2022survey}.
More sophisticated attention schemes, such as transformer-based models, can potentially capture longer-range dependencies and achieve even better performance in transfer tasks. Multi-head and hierarchical attention architectures are also worth exploring.
Many medical datasets lack comprehensive annotation. Attention mechanisms can be used to transfer knowledge from strongly annotated datasets to tasks with weak labels or small amounts of labeled data through techniques such as multiple-instance learning \cite{latif2023transformers}.
Electronic health records often contain multimodal data such as images, text reports, and genomic analyses. Attention provides a principled way to fuse heterogeneous data sources into unified predictive models in a transfer-learning setting \cite{shaik2023survey}.

\section{Conclusion}

This review has highlighted the significant potential and challenges of federated learning (FL) and transfer learning (TL) in advancing cancer detection through medical image analysis. FL's privacy-preserving capabilities allow for collaborative model training across multiple institutions without compromising patient data, while TL leverages pre-existing models to improve performance even with limited labeled data. Together, they represent a promising direction for overcoming current limitations in medical image analysis, particularly in cancer detection.
However, challenges such as domain shift, computational and hardware limitations, data privacy, heterogeneity of cancer, and the limited availability of annotated medical data persist. These challenges highlight the need for ongoing research to refine these methodologies and ensure their applicability in real-world clinical settings.

Future directions in this field should focus on enhancing models' ability to handle non-IID and heterogeneous data more effectively; improving the  accuracy and general applicability of FL and TL in diverse medical contexts; and developing strategies for collaborative FL to further enhance privacy preservation while leveraging the collective strength of distributed data; exploring decentralized training paradigms to mitigate central points of failure and further enhance privacy and security; advancing domain adaptation techniques within TL to better manage the domain shift and improve model performance across different medical imaging tasks; leveraging multimodal TL to integrate diverse data types and sources for a more holistic understanding and detection of cancer; and incorporating attention mechanisms in TL to improve model interpretability and efficacy, particularly in complex cancer detection tasks.

As the fields of FL and TL continue to evolve, their integration into clinical workflows promises not only to enhance cancer detection accuracy but also to democratize access to advanced diagnostic tools. This will require a concerted effort from researchers, clinicians, and policymakers to navigate the ethical, technical, and operational challenges ahead. Ultimately, the goal is to harness the power of collaborative and intelligent computing to save lives and improve health outcomes globally.


\section*{Data availability}
Data will be made available on request.

\section*{Conflict of Interest}
The authors declare no conflicts of~interest.

\end{document}